\documentclass{article}

\usepackage[numbers]{natbib}
\usepackage[table]{xcolor}
\definecolor{cvprblue}{rgb}{0.21,0.49,0.74}
\usepackage[pagebackref,breaklinks,colorlinks=true,linkcolor=magenta,citecolor=cvprblue]{hyperref}

\usepackage[preprint]{neurips_2026}

\usepackage[utf8]{inputenc} %
\usepackage[T1]{fontenc}    %
\usepackage{hyperref}       %
\usepackage{url}            %
\usepackage{booktabs}       %
\usepackage{amsfonts}       %
\usepackage{nicefrac}       %
\usepackage{microtype}      %
\usepackage{xspace}
\usepackage{enumitem}
\usepackage{amsmath}
\usepackage{amssymb}
\usepackage{graphicx}
\usepackage{subcaption}

\newcommand{\eg}{\textit{e.g.}}
\newcommand{\ie}{\textit{i.e.}}
\newcommand{\myparagraph}[1]{\noindent\textbf{#1}}
\newcommand{\mycaption}[2]{\caption{\textbf{#1.}~#2}}
\newcommand\ourmodel{\emph{CineOrchestra}\xspace}
\newcommand\ourbench{\emph{CineBench}\xspace}
\newcommand\ourbenchsyn{\emph{CineBenchSyn}\xspace}

\title{\ourmodel: Unified Entity-Centric Conditioning for Cinematic Video Generation}

\author{%
  Sharath Girish\textsuperscript{1,$\ast$} \quad 
  Tsai\mbox{-}Shien Chen\textsuperscript{1,2,$\ast$} \quad 
  Zhikang Dong\textsuperscript{1} \quad 
  Mukesh Singhal\textsuperscript{2} \\
  \vspace{5pt} \\
  \textbf{Hao Chen\textsuperscript{1}} \quad 
  \textbf{Sergey Tulyakov\textsuperscript{1}} \quad 
  \textbf{Aliaksandr Siarohin\textsuperscript{1}} \\
  \vspace{5pt} \\
  \textsuperscript{1}Snap Inc. \quad \textsuperscript{2}UC Merced \\ \\
  Project page: \href{https://snap-research.github.io/CineOrchestra/}{snap-research.github.io/CineOrchestra/}
}

\begin{document}

\maketitle

\renewcommand{\thefootnote}{\fnsymbol{footnote}}
\footnotetext[1]{Equal contribution.}
\renewcommand{\thefootnote}{\arabic{footnote}}

\begin{figure}[h]
    \centering
    \includegraphics[width=\linewidth]{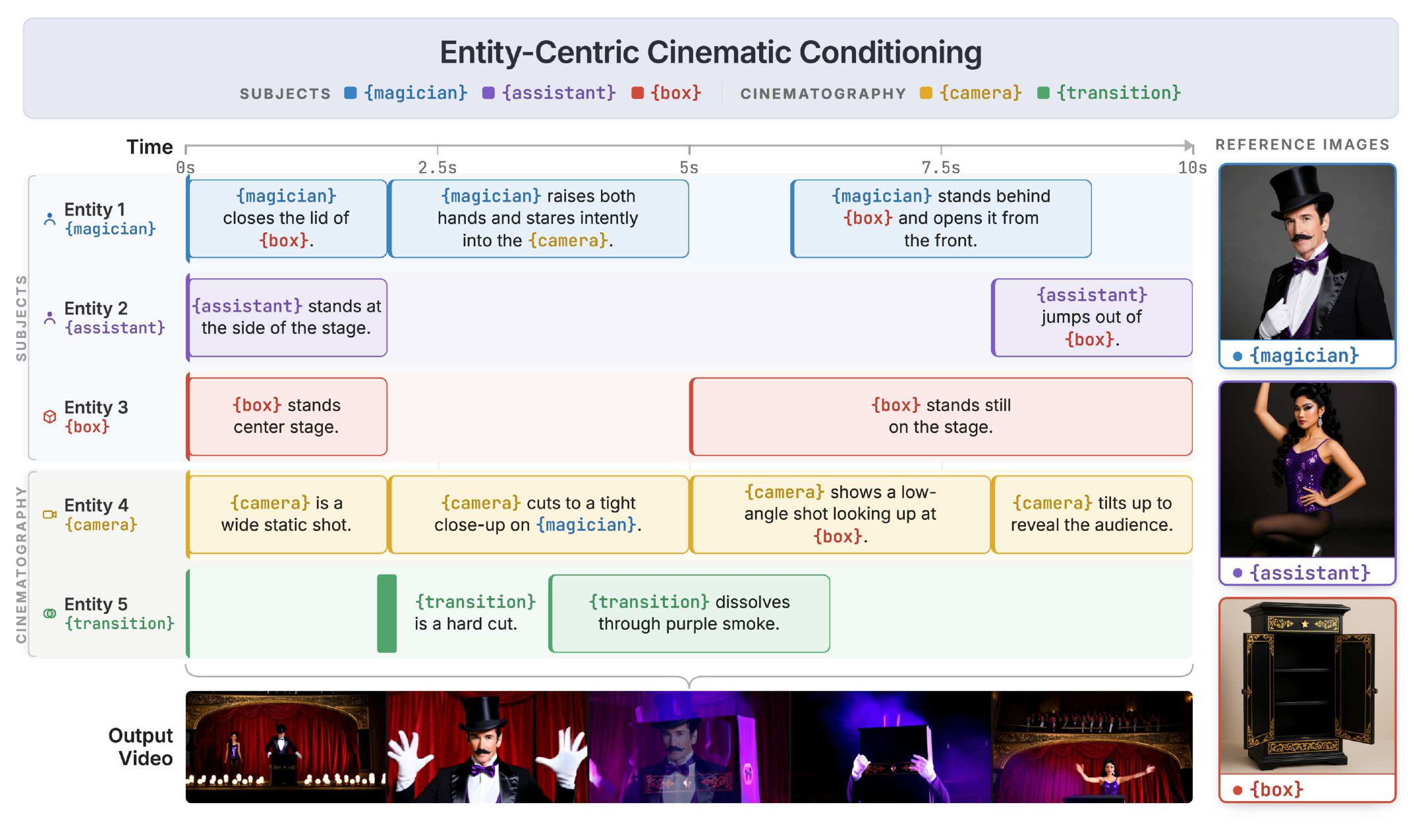}
    \mycaption{\ourmodel generates cinematic scenes from unified conditioning}{
        Our entity-centric conditioning represents every cinematic element (\textit{visual subjects}, \textit{camera}, and \textit{shot transitions}) as a unified, timestamped expression (top), with optional reference images (top right).
        It enables \ourmodel to generate cinematic frames in a single forward pass, jointly realizing multi-subject personalization, multi-event timing, multi-shot composition, and camera control (bottom).
    }
    \label{fig:teaser}
\end{figure}

\begin{abstract}

Cinematic video depicts multiple subjects acting or interacting at specific moments, captured with deliberate camera movement, and stitched together by shot transitions.
Together, these elements demand a level of fine-grained control beyond current text-to-video models.
Existing work addresses each axis in isolation: multi-subject personalization, temporal control, multi-shot synthesis, or camera control; no prior framework jointly integrates all four.
We present \ourmodel, a unified video diffusion model that controls subjects, events, cameras, and shot transitions simultaneously.
Our key insight is that these heterogeneous cinematic elements share a fundamental structure: each is an entity acting over a specific temporal interval, which can therefore all be expressed through one shared structure of  entity-centric conditioning primitives, augmented with reference images for visual entities. 
This formulation reduces the architectural challenge to a single positional encoding problem, which we solve with two parameter-free coordinated rotary embeddings: \textit{(i)} an interval-sampled temporal RoPE that yields consistent attention behavior across events of dramatically varying duration,  and \textit{(ii)} a 2D entity-temporal cross-attention RoPE that disambiguates per-entity conditions and routes each to its corresponding spatiotemporal region.
On two new benchmarks, \ourmodel outperforms six per-axis specialists on dense caption following and shot-transition timing, with consistent gains in a pairwise user study and component ablations.

\end{abstract}

\section{Introduction}
\label{sec:introduction}

A scene from a movie is rarely a \textit{single static shot} of a \textit{single subject} performing a \textit{single action}. Instead, several characters and objects co-exist, each acting or interacting with others at specific moments, captured by a camera that moves with intent, and stitched together by transitions between distinct shots. This makes cinematic video generation an inherently compositional problem that remains out of reach for current text-to-video models~\cite{sora,wan,hunyuanvideo,cogvideox,veo,moviegen,ltx}, which typically condition on a single global caption and produce a single shot.

Recent work has begun to decompose this monolithic conditioning, with each line targeting one axis of cinematic composition:
\textit{(i)} multi-reference personalization models~\cite{video_alchemist,phantom,skyreels,magref} compose co-existing entities;
\textit{(ii)} temporal control methods~\cite{mint,alchemint} manipulate when each event happens;
\textit{(iii)} multi-shot synthesis methods~\cite{cinetrans,echoshot,multishotmaster,shotstream} produce several connected shots; and
\textit{(iv)} camera-conditioning models~\cite{cameractrl,motionctrl} allow user-specified viewpoint motion.
Each line, however, uses bespoke architectures trained on disjoint data; no existing model jointly ingests \textit{subjects}, \textit{events}, \textit{shot transitions}, and \textit{camera} and produces a coherent cinematic video scene.

We present \ourmodel, a cinematic video generation framework that controls all four axes within a single model. Our key insight is that these heterogeneous elements share a common structure: each is an entity acting over a temporal interval. We therefore propose an \textit{entity-centric cinematic conditioning} that expresses every cinematic element as a structured tuple of (start\_time, end\_time, prompt), as illustrated in \cref{fig:teaser}. Each character and object receives a unique tag (\eg, \texttt{\{man\}}, \texttt{\{car\}}), a global identity description, a set of event-level dense descriptions (\eg, ``[0.1s\,--\,2.3s] \texttt{\{man\}} jumps into \texttt{\{car\}}''), and an optional reference image to provide identity details. Crucially, we extend the same structure to cinematography through two special tags, \texttt{\{camera\}} and \texttt{\{transition\}}, which carry only event-level descriptions (\eg, ``[6.3s\,--\,6.4s] \texttt{\{transition\}} shows a hard cut''; ``[0.0s\,--\,10.0s] \texttt{\{camera\}} pans left across \texttt{\{car\}}''). The same representation thus captures the full cinematic structure with no modality-specific design.

This unification shifts the entire technical challenge onto a single problem: positional encoding~\cite{transformer,rope} in our video DiT backbone~\cite{dit,snapvideo}. A single clip may carry many reference images and dozens of overlapping events whose durations span a dramatically wide range from 0.1s hard cuts to 10s sustained camera moves, which standard fixed-cadence temporal RoPE~\cite{mint} cannot fairly represent. We therefore introduce two coordinated RoPE designs: a \textit{interval-sampled temporal RoPE} that evenly samples position within each event's interval and rescales them for a duration-invariant similarity peak, and a \textit{2D entity-temporal cross-attention RoPE} that disambiguates entity tokens and routes each conditioning element to its corresponding spatiotemporal region.

On two newly proposed benchmarks, \ourbench and \ourbenchsyn, \ourmodel outperforms six per-axis specialists on dense caption following and shot-transition timing, confirmed by a pairwise user study and ablations of each component. Our contributions can be summarized as follows:
\begin{itemize}[leftmargin=*,itemsep=0pt,topsep=0pt,parsep=0pt,partopsep=0pt]
    \item We present \ourmodel, the first framework for joint cinematic video generation across subjects, events, camera, and shot transitions, built on a unified entity-centric conditioning primitive.
    \item We introduce two parameter-free coordinated RoPEs that handle variable-duration events, disambiguate per-entity conditions, and route each to its target spatiotemporal region.
    \item We introduce two benchmarks, \ourbench and \ourbenchsyn, on which \ourmodel outperforms six per-axis specialists in both automatic metrics and a pairwise user study.
\end{itemize}

\section{Related Work}
\label{sec:related_work}

Generating cinematic video scenes requires simultaneous control over character identity, narrative timing, shot structure, and camera movement, where prior work addresses each axis in isolation.

\myparagraph{Video diffusion models.}
Diffusion models~\cite{dpm,ncsn,ddpm,ldm,vdm} have driven tremendous progress in text-to-video and image-to-video generation through training on Internet-scale data~\cite{imagen_video,svd,mcdiff,make_a_video,hunyuanvideo,panda}.
While earlier approaches adopt U-Net~\cite{unet} as the denoising backbone~\cite{align_your_latents,animatediff,videocrafter1,videocrafter2}, recent models have switched to diffusion transformers~\cite{dit} (DiTs) whose scalability better handles high-resolution, long, and visually complex videos~\cite{sora,wan,cogvideox,WALT,moviegen,veo,ltx}.
\ourmodel builds on a video DiT~\cite{snapvideo} but further extends it to enable cinematic compositional control.

\myparagraph{Multi-subject personalization.}
Subject personalization began with optimization-based methods~\cite{dreambooth,textual_inversion,multi_concept_customization} and matured into feed-forward reference-image conditioning~\cite{ip_adapter,instant_id,instant_booth,canvas2image,layercomposer}.
Recent work extends the idea to multiple subjects~\cite{video_alchemist,phantom,concept_master,magref,skyreels,tora2,vimi,omni_attribute}, typically by injecting reference-image tokens through attention operations~\cite{transformer} to render several identities in one coherent video.
While these methods ground multiple identities, they treat the clip as a single global event with one caption and cannot express the time-localized scripts that cinematic video scenes require.

\myparagraph{Multi-event temporal control.}
Videos add a temporal dimension over frames, motivating work that decomposes a video into time-localized segments.
\textit{Multi-shot} methods~\cite{cinetrans,echoshot,multishotmaster,shotstream} synthesize several connected shots separated by hard cuts, but each per-shot caption is monolithic, leaving no dense intra-shot timing.
\textit{Continuous-frame} methods~\cite{mint,alchemint} address this by generating a single video from dense, time-stamped captions with a temporal RoPE that biases attention toward each caption's annotated interval, and additionally accept a separate scene-cut conditioning track for shot transitions. However, this track encodes only cut \textit{timing}, with no natural-language description of the transition (e.g., dissolve, wipe, fade).
In contrast, \ourmodel unifies \textit{both} under one entity-centric primitive: dense captions handle intra-shot timing, while a \texttt{\{transition\}} tag carries a dense description of inter-shot transitions and composes with other entity tags, all in one forward pass.

\myparagraph{Cinematography conditioning.}
Camera motion is a primary expressive tool in filmmaking, making viewpoint control an important axis of video generation.
Existing methods condition generation on explicit geometric signals, such as Plücker-embedded camera trajectories~\cite{cameractrl,vd3d,cami2v,ac3d}, joint camera-and-object pose sequences~\cite{motionctrl}, or novel-view re-rendering of a source video~\cite{recammaster}.
However, such inputs demands either specialized capture rigs, 3D reconstruction pipelines, or manual pose authoring. \ourmodel instead conditions camera behavior through time-stamped natural-language descriptors (\eg, ``[1.5s\,--\,6.7s] \texttt{\{camera\}} pushes in slowly''; ``[6.7s\,--\,9.0s] \texttt{\{camera\}} pans left to reveal \texttt{\{man\}}''), retaining directorial control without requiring any explicit pose inputs.

\section{\ourmodel}
\label{sec:method}

\ourmodel reformulates cinematic video generation as conditioning a single video diffusion transformer~\cite{dit,snapvideo} (DiT) on a unified, entity-centric description of cinematic structure. \cref{sec:method:architecture} presents the overall architecture, which inherits the standard self-attention then cross-attention layout in recent multi-reference video diffusion models~\cite{video_alchemist,phantom,alchemint,skyreels,magref}. Our key technical novelty lies in the conditioning primitive (\cref{sec:method:conditioning}) and the rotary positional embeddings (\cref{sec:method:event_rope,sec:method:self_attn,sec:method:cross_attn}). Lastly, \cref{sec:method:data} describes the data curation pipeline that supplies entity-centric annotations at scale.

\begin{figure}[t]
    \centering
    \includegraphics[width=.9\linewidth]{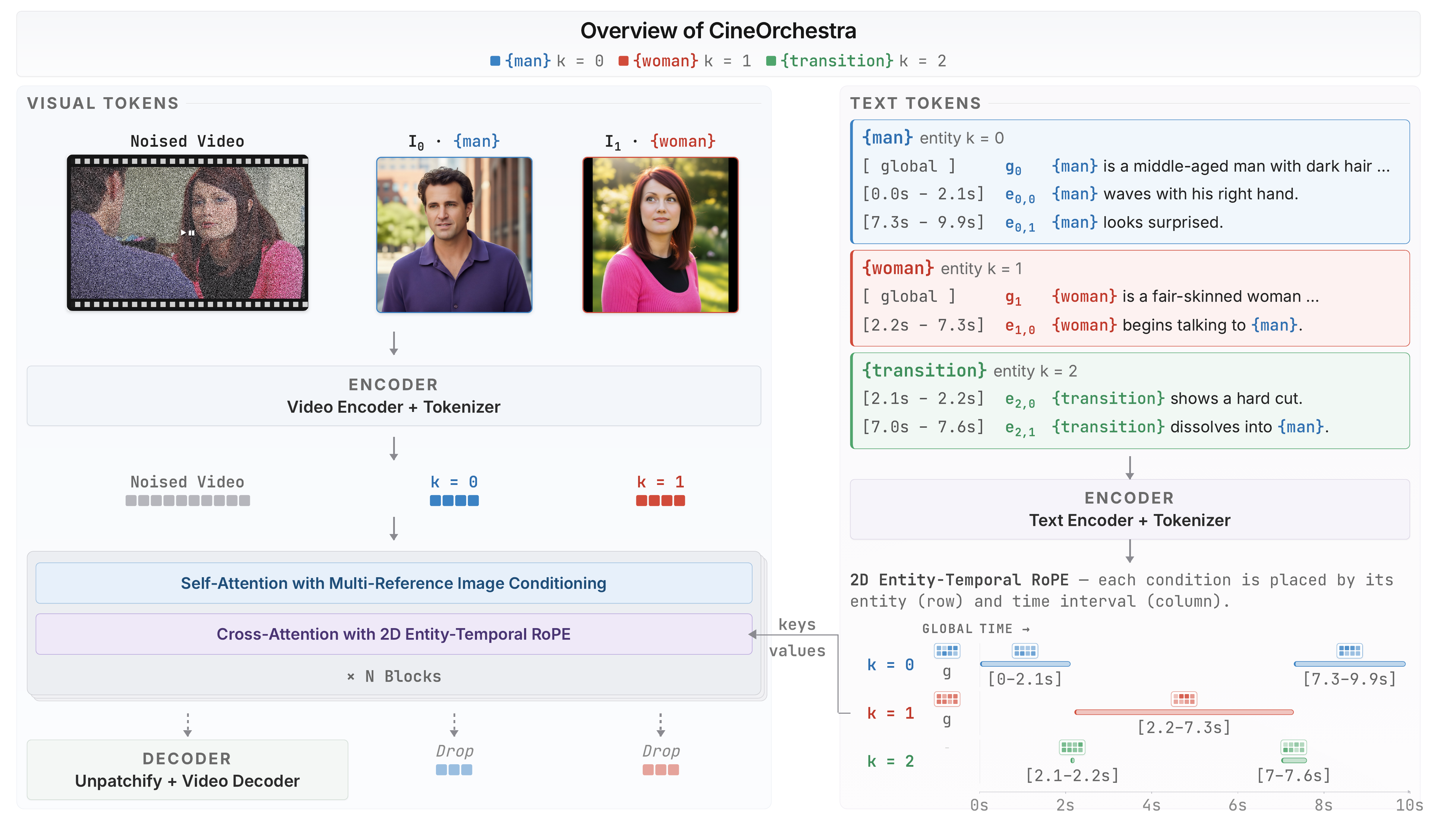}
    \mycaption{Overview of \ourmodel}{
        Each entity $k$ is represented by a reference image $\mathbf{I}_k$, a global description $g_k$, and a set of event-level dense descriptions $\{(t^s_{k,j}, t^e_{k,j}, e_{k,j})\}$ pairing temporal intervals with prompts. Reference image tokens are concatenated to the video tokens for full self-attention, while all text tokens are consumed via cross-attention. Two coordinated RoPEs, interval-sampled temporal RoPE (\cref{sec:method:event_rope}) and 2D entity-temporal cross-attention RoPE (\cref{sec:method:cross_attn}), disambiguate per-entity conditions and route each to its target spatiotemporal region.
    }
    \label{fig:architecture}
\end{figure}

\subsection{Architecture Overview}
\label{sec:method:architecture}

As illustrated in \cref{fig:architecture}, we build on a video DiT~\cite{snapvideo} that operates on spatiotemporally-patchified video tokens $\mathbf{V}$ produced by a video variational autoencoder~\cite{vae} (VAE). To support multi-reference image conditioning, each input image is encoded by the same VAE as a single-frame video, yielding per-entity tokens $\{\mathbf{I}_k\}_{k=1}^{K}$ at the same spatial resolution as $\mathbf{V}$, where $K$ is the number of entities in the clip. Following Video Alchemist~\cite{video_alchemist}, we apply random augmentations to each reference image before encoding to mitigate copy-paste artifacts. We then append these tokens to the video sequence and apply full self-attention over the concatenated stream $[\mathbf{V}; \mathbf{I}_1; \dots; \mathbf{I}_K]$.

To support text conditioning, each text prompt (including a global identity caption per entity and a dense caption per event) is independently encoded with T5~\cite{t5} and concatenated into a single key-value bank. Each transformer block then applies cross-attention where $[\mathbf{V}; \mathbf{I}_1; \dots; \mathbf{I}_K]$ serves as queries and this bank serves as keys and values.

Notably, our framework adds no learnable parameters to the underlying video DiT: every component reuses an existing module, and our two coordinated RoPEs in \cref{sec:method:event_rope,sec:method:cross_attn} are parameter-free modifications to positional encoding. \cref{app:implementation:architecture} completes the architectural specifications.

\subsection{Entity-Centric Cinematic Conditioning}
\label{sec:method:conditioning}

We represent every cinematic element through a single entity-grounded primitive. Each entity $k \in \{1, \dots, K\}$ is associated with the following fields:
\begin{itemize}[leftmargin=*,itemsep=0pt,topsep=0pt,parsep=0pt,partopsep=0pt]
    \item A unique tag $\tau_k$ that serves as a stable referent across all captions (\eg, \texttt{\{man\_old\}}, \texttt{\{car\_red\}}).
    \item An optional reference image $\mathbf{I}_k$ that faithfully encodes visual identity.
    \item An optional global identity description $g_k$ (\eg, ``\texttt{\{man\_old\}} is an 80-year-old man with short white hair wearing a black coat'').
    \item A set of event-level dense descriptions $\mathcal{E}_k = \{\dots, (t^s_{k,j}, t^e_{k,j}, e_{k,j}), \dots\}$, where each tuple pairs a temporal interval with a prompt (\eg, ``[0.1s\,--\,1.3s] \texttt{\{man\_old\}} raises his hand''; ``[1.3s\,--\,2.9s] \texttt{\{man\_old\}} then waves his hand''). The first token in the prompt identifies its primary entity.
\end{itemize}

Our conditioning primitive features two advantages. First, movie-level scenes routinely involve complex interactions between entities. Since every $\tau_k$ is a stable referent, dense descriptions can express such interactions unambiguously by including multiple tags (\eg, ``[2.9s\,--\,4.2s] \texttt{\{man\_old\}} opens the door of \texttt{\{car\_red\}}'').
 
Second, in addition to visual subjects, cinematographic elements such as camera movement and shot transitions are also essential to a cinematic scene. Our primitive can be naturally extended to these via two reserved tags, \texttt{\{camera\}} and \texttt{\{transition\}}, which carry only event-level dense descriptions. More importantly, they are directly composable with visual subjects by referencing their tags (\eg, ``[0.0s\,--\,7.8s] \texttt{\{camera\}} pans left across \texttt{\{car\_red\}}''). Furthermore, unlike prior multi-shot methods~\cite{cinetrans,echoshot,multishotmaster,shotstream} that implicitly assume instantaneous cuts, our dense descriptions enable diverse transition types (\eg, ``[5.7s\,--\,6.2s] \texttt{\{transition\}} fades to black'').

\begin{figure}[t]
    \centering
    \begin{subfigure}[t]{0.38\linewidth}
        \centering
        \caption{Interval-Sampled Temporal RoPE.}
        \includegraphics[width=\linewidth]{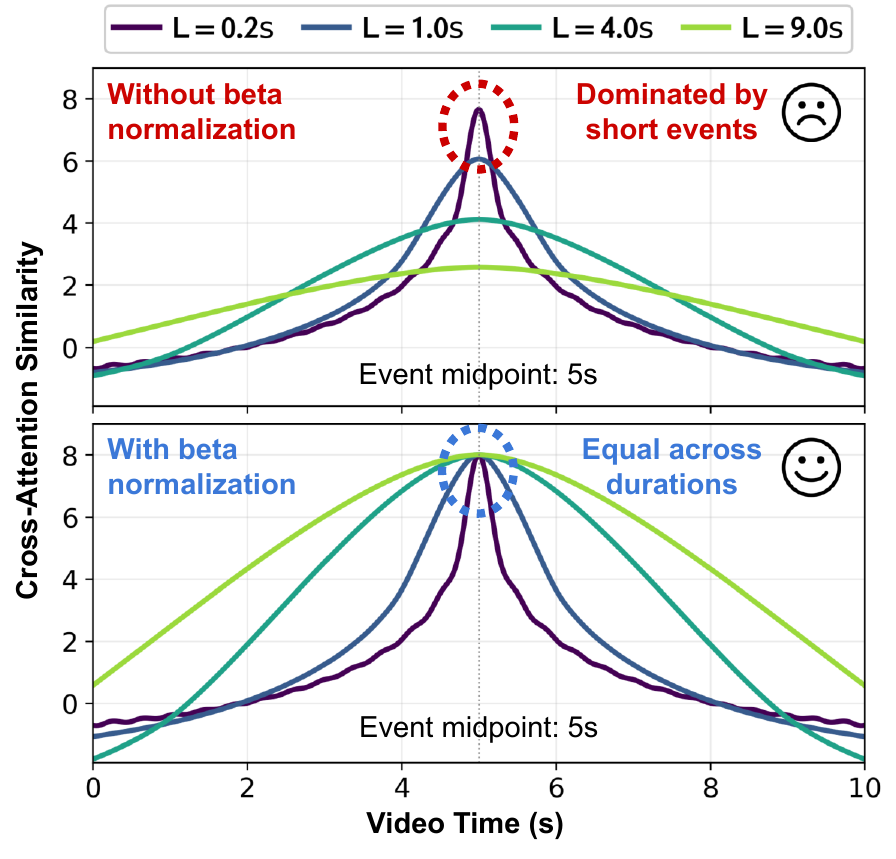}
        \label{fig:event_rope}
    \end{subfigure}
    \hfill
    \begin{subfigure}[t]{0.60\linewidth}
        \centering
        \caption{2D Entity-Temporal Cross-Attention RoPE.}
        \includegraphics[width=\linewidth]{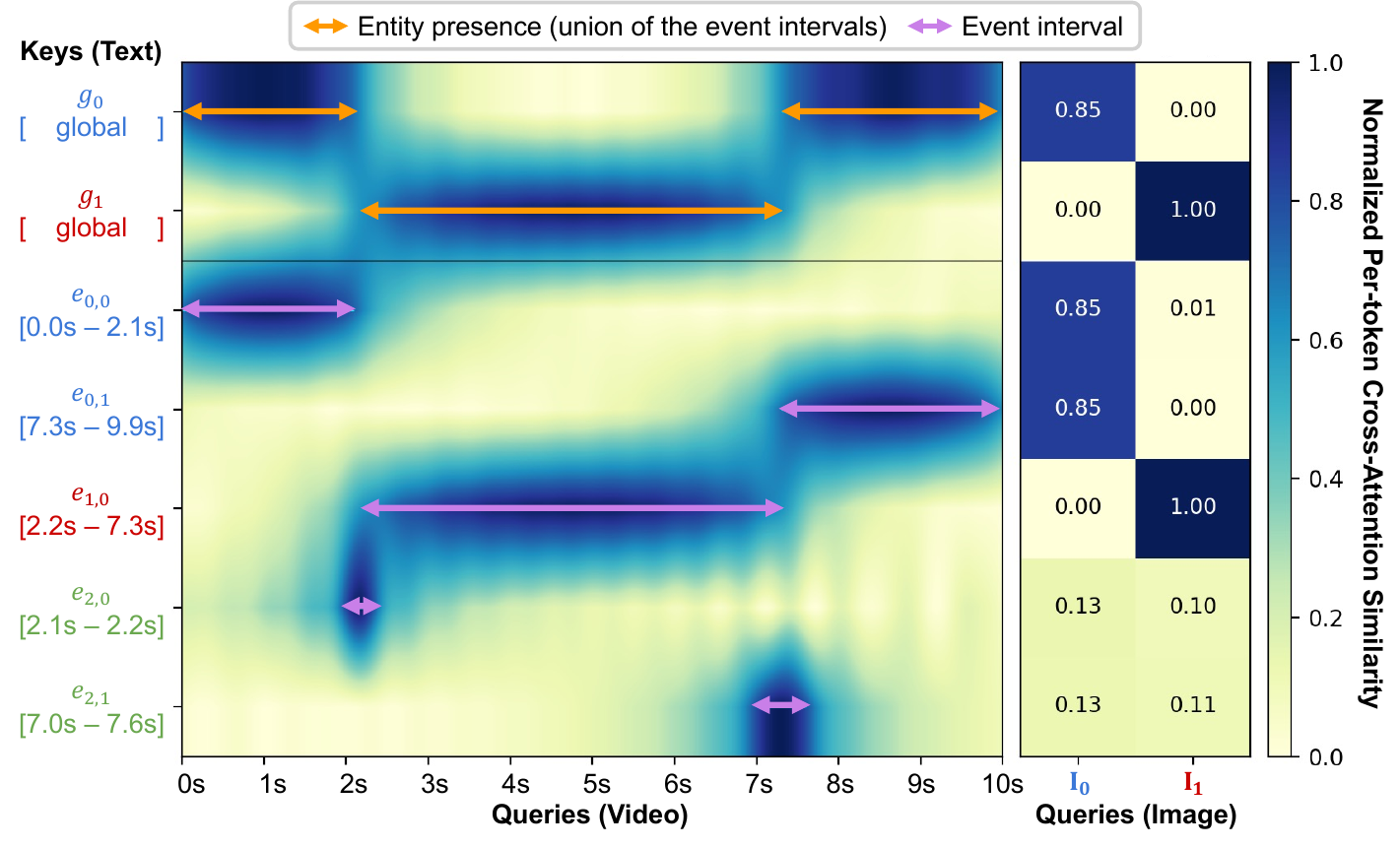}
        \label{fig:2d_rope}
    \end{subfigure}
    \mycaption{Two coordinated RoPE designs}{
        (a) Similarity between a video token and an event token across event durations $L$. Our $\beta(L)$ rescaling produces duration-invariant peaks.
        (b) Cross-attention similarity between video/image queries and global/dense-description keys under \cref{tab:2d_rope}'s coordinates. Sharp peaks emerge only where query and key share the same entity and overlap in time, jointly achieving entity \textit{disambiguation} and temporal \textit{routing}.
    }
    \label{fig:rope}
\end{figure}

\subsection{Interval-Sampled Temporal RoPE}
\label{sec:method:event_rope}
 
Durations of cinematic events $L$ could span a dramatically wide range: a hard cut may last 0.1s while a sustained pan or character action may last 10s. Prior multi-event temporal control methods~\cite{mint} encode each event with a \textit{fixed-cadence} temporal RoPE at a constant temporal interval $\Delta t$ (\eg, per frame or per second), which suffers two failure modes:
\begin{itemize}[leftmargin=*,itemsep=0pt,topsep=0pt,parsep=0pt,partopsep=0pt]
    \item Sub-cadence events falling below the encoding resolution ($L < \Delta t$) become indistinguishable.
    \item The attention similarity against video tokens depends on $L$, biasing attention toward shorter events.
\end{itemize}

We address both issues through the proposed \textit{interval-sampled temporal RoPE}. Formally, let $R(t) \in \mathbb{R}^{d_{\text{rope}} \times d_{\text{rope}}}$ denote the standard RoPE rotation~\cite{rope} at temporal coordinate $t$, where $d_{\text{rope}}$ is the per-head channel dimension. For an event during $[t_s, t_e]$, we sample $N=16$ positions evenly \emph{within} the interval and define its temporal positional encoding as the rescaled average:
\begin{equation}
    \mathbf{P}^{\text{event}}(t_s, t_e) \;=\; \beta(L) \cdot \frac{1}{N}\sum_{i=0}^{N-1} R\!\left(t_s + \tfrac{i}{N-1}(t_e - t_s)\right),
    \label{eq:event_rope}
\end{equation}
where $\beta(L)$ is a duration-dependent scalar specified below.
By placing $N$ positions \emph{within} the interval rather than sampling at a fixed cadence $\Delta t$, $\mathbf{P}^{\text{event}}$ resolves the sub-cadence failure mode and jointly captures the event's start, end, and duration in a single positional encoding.
 
\myparagraph{Similarity-peak normalization.}
Setting $\beta\equiv 1$, however, leaves the second failure mode. The dot product between an event token and a video token undergoes \textit{phase cancellation} across the $N$ samples, with magnitude decaying as $\mathrm{sinc}(\theta_n L/2)$ for each RoPE frequency $\theta_n$. Consequently, wider intervals receive smaller-magnitude peaks than narrower ones (see \cref{fig:rope}(a) for visualization and \cref{app:beta} for the full expression). To address this, we introduce a duration-dependent scalar
\begin{equation}
    \beta(L) \;=\; \sqrt{d_{\text{rope}}} \;\Big/\; \big\|\tfrac{1}{N}\textstyle\sum^{N-1}_{i=0} R(t_i)\big\|_F,
    \label{eq:beta_closed_form}
\end{equation}
so that the Frobenius norm $\|\mathbf{P}^{\text{event}}(t_s, t_e)\|_F$ is duration-invariant.
\cref{app:beta} derives the closed form of $\beta(L)$ and establishes three key properties: it \textit{(i)} reduces to standard RoPE at $L = 0$, \textit{(ii)} remains well-bounded for all event durations, and \textit{(iii)} yields an approximately duration-invariant peak similarity.
\cref{fig:rope}(a) visualize its effect by comparing the cross-attenton similarity (between a video token and an event token) without and with normalization. \cref{sec:exp:ablation} further validates the design empirically.

\subsection{Self-Attention: Multi-Reference Image Conditioning}
\label{sec:method:self_attn}
 
Self-attention operates over the concatenated stream $[\mathbf{V}; \mathbf{I}_1; \dots; \mathbf{I}_K]$, making video and image tokens share the same positional space. In such a case, applying vanilla 3D RoPE could place each $\mathbf{I}_k$ at coordinates that collide with arbitrary regions of $\mathbf{V}$, creating ambiguity between video and image tokens. We address this with separate spatial and temporal RoPE designs for image tokens.
 
\myparagraph{Spatial RoPE for image tokens.}
Following Qwen-Image~\cite{qwen_image_edit}, we place each reference image on its own diagonal block of the spatial RoPE plane: video occupies $[0, H) \times [0, W)$, while image $k$ occupies $[kH, (k+1)H) \times [kW, (k+1)W)$, disjoint from the video and from every other image.
 
\myparagraph{Temporal RoPE for image tokens.}
Visual identity is meaningful only when an entity is on screen. For instance, the reference image of \texttt{\{man\_old\}} should influence only the frames in which he appears. We therefore anchor each reference image to the temporal extent of its events. That is, $\mathbf{I}_k$ inherits a temporal position equal to the mean of its event-level temporal RoPEs
\begin{equation}
    \mathbf{P}^{\text{image}}(k) \;=\; \frac{1}{|\mathcal{E}_k|} \sum_{(t_s, t_e, e) \in \mathcal{E}_k} \mathbf{P}^{\text{event}}(t_s, t_e),
    \label{eq:image_rope}
\end{equation}
which biases self-attention to route identity into the correct temporal region of the video.

\begin{table}[t]
    \centering
    \mycaption{Coordinates of 2D entity-temporal RoPE per cross-attention token type}{
        $\mathbf{P}^{\text{event}}$ and $\mathbf{P}^{\text{image}}$ are defined in \cref{eq:event_rope,eq:image_rope}, respectively.
        The entity-index axis \textit{disambiguates} per-entity conditions, while the temporal axis \textit{routes} each to its target spatiotemporal region.
    }
    \vspace{+2mm}
    \label{tab:2d_rope}
    \small
    \setlength{\tabcolsep}{10pt}
    \begin{tabular}{lll}
        \toprule
        \textbf{Token type} & \textbf{Entity-index axis} & \textbf{Temporal axis} \\
        \midrule
        Video token $\mathbf{V}$ (query)       & $\frac{1}{K}\sum_{k=1}^{K} R_{\text{entity}}(k)$  & 3D-RoPE temporal coordinate \\
        Reference image $\mathbf{I}_k$ (query) & $R_{\text{entity}}(k)$                            & $\mathbf{P}^{\text{image}}(k)$ \\
        Global description $g_k$ (key)         & $R_{\text{entity}}(k)$                            & $\mathbf{P}^{\text{image}}(k)$ \\
        Dense description $e_{k,j}$ (key)      & $R_{\text{entity}}(k)$                            & $\mathbf{P}^{\text{event}}(t_s^{k,j}, t_e^{k,j})$ \\
        \bottomrule
    \end{tabular}
\end{table}

\subsection{Cross-Attention: 2D Entity-Temporal RoPE}
\label{sec:method:cross_attn}

Cross-attention injects text conditioning into the video and image queries. A single clip may present $K$ reference images, $K$ global descriptions, and $\sum_{k=1}^{K}|\mathcal{E}_k|$ event-level dense descriptions spanning visual subjects, camera, and shot transition. With so many parallel and heterogeneous conditioning, the positional encoding must satisfy two requirements: \textit{(i) disambiguation}: image and text tokens for different entities or events must remain positionally distinct; \textit{(ii) routing}: each conditioning element must bias attention toward the entity and frames it describes.

\myparagraph{2D entity-temporal RoPE.}
We meet \textit{both} by introducing a 2D RoPE with \textit{(i)} an \textit{entity-index axis} that separates entities to enforce disambiguation and \textit{(ii)} a \textit{temporal axis} that aligns each conditioning element with its target frames for routing.
Specifically, the entity-index axis is encoded by a standard RoPE rotation $R_{\text{entity}}(k) \in \mathbb{R}^{d_{\text{entity}} \times d_{\text{entity}}}$ at integer index $k$, with one slot per entity.
The temporal axis is encoded by the interval-sampled temporal RoPE of \cref{sec:method:event_rope}.
Following the standard multi-axis RoPE construction~\cite{rope}, the per-head channel dimension is partitioned into two disjoint groups, with each axis's rotation applied to its own group, so similarity peaks only when both axes are aligned.

\myparagraph{Per-token coordinates.}
Given the two axes, \cref{tab:2d_rope} specifies the coordinates per token type:
\begin{itemize}[leftmargin=*,itemsep=0pt,topsep=0pt,parsep=0pt,partopsep=0pt]
    \item Video tokens $\mathbf{V}$ (query) are averaged over all $R_{\text{entity}}(k)$ on the entity-index axis, making them equally receptive to the conditioning of every entity present in the video.
    \item Reference image $\mathbf{I}_k$ (query) and global description $g_k$ (key) share the same coordinates since both encode entity-level rather than event-specific information. We anchor them at the entity-averaged temporal position $\mathbf{P}^{\text{image}}(k)$ to spread their influence softly whenever entity $k$ is on screen.
    \item Dense description $e_{k,j}$ (key) instead use the precise $\mathbf{P}^{\text{event}}(e_{k,j})$, producing a sharp similarity peak within their event interval and thereby satisfying the routing requirement.
\end{itemize}

This assignment delivers both \textit{disambiguation} and \textit{routing} without any auxiliary mask. The same coordinates apply uniformly to visual subjects, camera, and shot transition, so the unified primitive of \cref{sec:method:conditioning} is learned end-to-end with no entity-specific modules and no added parameters. \cref{fig:rope}(b) illustrates the 2D layout, and \cref{sec:exp:ablation} ablates each axis.

\subsection{Training Data Curation}
\label{sec:method:data}

We construct a dataset of one-minute chunks cropped from licensed movies and TV shows.
Within such a window, the cast and setting stay consistent, forming a closed entity set for reliable annotation.

\myparagraph{Entity-centric annotation.}
For each chunk, we issue a single structured-output query to Gemini 2.5~\cite{gemini} that populates every field of the primitive in \cref{sec:method:conditioning} and jointly extracts visual entities, camera, and shot transitions in one pass.

\myparagraph{Reference image with appearance augmentation.}
For each visual entity, we crop the entity from event middle frames using Gemini 3~\cite{gemini} bounding boxes, filter them by CLIP-T~\cite{clip} similarity to the entity's global description, and randomly sample two of the top matches. We then pass these crops to Qwen-Image-Edit~\cite{qwen_image_edit} to synthesize a new view of the same identity under a different pose, lighting, and facial expression.
Unlike the photometric and geometric augmentations of Video Alchemist~\cite{video_alchemist}, this step varies appearance while preserving identity, mitigating copy-paste artifacts during training.

\begin{table}[t]
    \centering
    \mycaption{Quantitative comparison of cinematic conditioning on \ourbench}{
        We compare \ourmodel against specialists: \textit{(i)} multi-reference personalization (top) and \textit{(ii)} multi-shot synthesis (middle).
        We report Masked DINO for subject identity consistency, ViCLIP for dense caption following, and Qwen-VL recall for shot-transition timing.
        Best in \textbf{bold}, second-best \underline{underlined}.
    }
    \vspace{+2mm}
    \label{tab:quantitative_real}
    \footnotesize
    \setlength{\tabcolsep}{4.5pt}
    \begin{tabular}{l|c|c|cccc|c}
        \toprule
        & \multicolumn{1}{c|}{\textbf{Subject ID}} & \multicolumn{1}{c|}{\textbf{Global Caption}} & \multicolumn{4}{c|}{\textbf{Dense Caption Following} (ViCLIP$\uparrow$)}  & \textbf{Transition Timing}  \\
        \cmidrule(lr){2-2}\cmidrule(lr){3-3}\cmidrule(lr){4-7}\cmidrule(lr){8-8}
        \textbf{Method}                        & M-DINO$\uparrow$ & M-CLIP$\uparrow$  & Subject & Scene & Camera & Transition & Recall$\uparrow$ \\
        \midrule
        Phantom~\cite{phantom}                 & \textbf{0.509} & \underline{0.295} & 0.212 & 0.197 & \underline{0.178} & \underline{0.146} & \underline{0.431} \\
        VACE~\cite{vace}                       & 0.482 & \textbf{0.309} & \underline{0.219} & \textbf{0.214} & 0.171 & 0.136 & 0.340 \\
        \midrule
        CineTrans~\cite{cinetrans}             & 0.423 & 0.286 & 0.214 & \underline{0.208} & 0.168 & 0.124 & 0.129 \\
        EchoShot~\cite{echoshot}               & 0.399 & 0.286 & 0.202 & 0.193 & 0.164 & 0.129 & 0.094 \\
        MultiShotMaster~\cite{multishotmaster} & 0.383 & 0.286 & 0.207 & 0.207 & 0.172 & 0.126 & 0.343 \\
        ShotStream~\cite{shotstream}           & 0.391 & 0.289 & 0.214 & 0.197 & 0.172 & 0.141 & 0.267 \\
        \midrule
        \rowcolor{gray!12}
        \textbf{\ourmodel (ours)}              & \underline{0.502} & \underline{0.295} & \textbf{0.235} & \underline{0.208} & \textbf{0.193} & \textbf{0.150} & \textbf{0.486} \\
        \bottomrule
    \end{tabular}
\end{table}

\section{Experiments}
\label{sec:experiment}

We validate that \ourmodel jointly controls all four cinematic axes (\textit{subjects}, \textit{events}, \textit{camera}, \textit{shot transitions}) in a single model. \cref{sec:exp:comparison} compares against per-axis specialized baselines. \cref{sec:exp:ablation} ablates the proposed components of \cref{sec:method}. \cref{app:implementation} provides additional training and inference details.

\begin{figure*}[t]
    \centering
    \includegraphics[width=\linewidth]{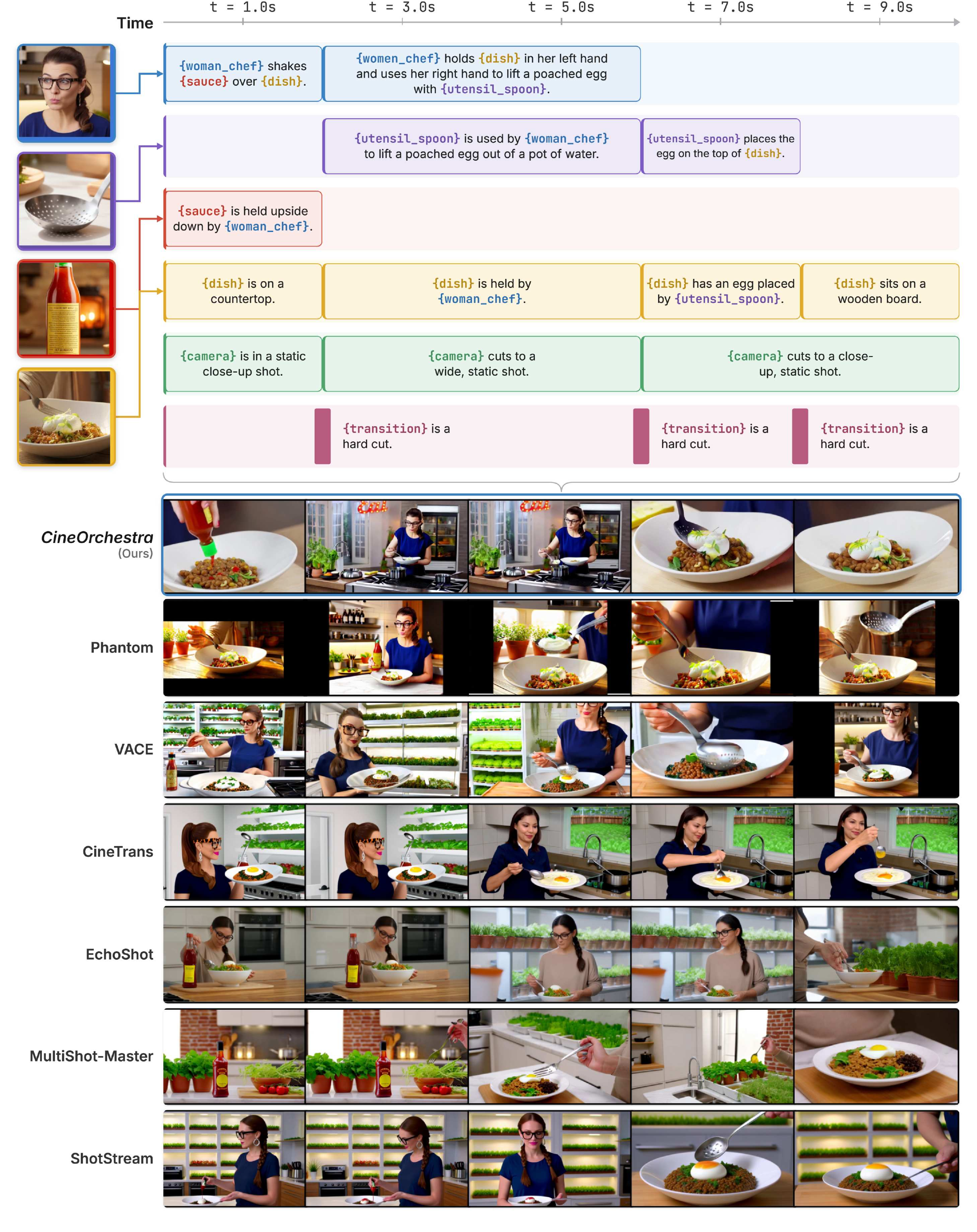}
    \mycaption{Qualitative comparison of cinematic conditioning on \ourbench}{
        Given the entity-centric conditioning (top), \ourmodel (top video row) simultaneously preserves all four subject identities, follows the dense per-entity timeline, and lands three hard cuts, outperforming all existing methods. More comparisons on \ourbenchsyn can be found in \cref{app:additional_results}.
    }
    \label{fig:qualitative_main}
\end{figure*}

\begin{figure}[t]
    \centering
    \includegraphics[width=\linewidth]{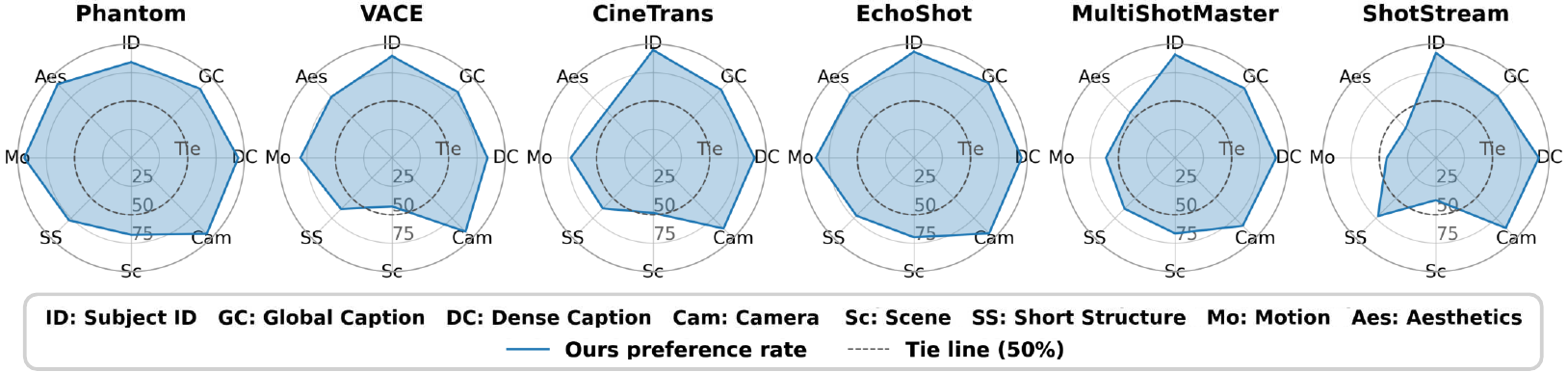}
    \mycaption{User study on \ourbench}{
        Pairwise preference of \ourmodel against six baselines on eight dimensions (one radar panel per baseline). Each axis reports $\mathrm{pref} = W / (W + L)$, the share of \textit{decisive} votes that favoured \ourmodel, where $W$ counts wins, $L$ counts losses, and ties are excluded. The dashed circle marks the $50\%$ tie line: points outside favour \ourmodel, points inside favour the baseline.
        \ourmodel wins on every entity-, text-, and structure-related axis against all six baselines, while perceptual axes (\textit{motion}, \textit{overall quality}, \textit{scene}) tie or favour the strongest perceptual baselines.
    }
    \label{fig:user_study}
\end{figure}

\subsection{Comparisons across Four Cinematic Axes}
\label{sec:exp:comparison}

\myparagraph{Baselines.}
Since no prior framework jointly handles all four axes, we compare against per-axis specialists from the two axes whose baselines accept entity-centric reference inputs:
\textit{(i) multi-reference personalization} from Phantom~\cite{phantom} and VACE~\cite{vace}, and
\textit{(ii) multi-shot synthesis} from CineTrans~\cite{cinetrans}, EchoShot~\cite{echoshot}, MultiShotMaster~\cite{multishotmaster}, and ShotStream~\cite{shotstream}.
Each baseline uses its official checkpoint, adapted to our entity-centric inputs as detailed in \cref{app:evaluation:baselines}.

\myparagraph{Benchmark datasets.}
We introduce two complementary benchmarks as no existing benchmark covers all four cinematic axes:
\textit{(i)} \ourbench ($512$ clips, $3.2k$ entities, $6.9k$ events, $1.5k$ reference images) contains movie and TV clips from titles unseen during training, annotated with our entity-centric primitive.
\textit{(ii)} \ourbenchsyn ($512$ clips, $3.3k$ entities, $6.4k$ events, $1.7k$ reference images) includes LLM-generated prompts and Qwen-Image-generated reference images targeting under-represented edge cases.
Dataset statistics are provided in \cref{app:implementation:dataset} and the curation of \ourbenchsyn in \cref{app:evaluation:benchsyn}.
We will release \ourbenchsyn for future research on cinematic video generation.

\myparagraph{Evaluation metrics.}
We report seven metrics to comprehensively evaluate all four axes (see the full definitions of each metric in \cref{app:evaluation:metrics}):
\begin{itemize}[leftmargin=*,itemsep=0pt,topsep=0pt,parsep=0pt,partopsep=0pt]
    \item \textit{Subject identity consistency}: following Video Alchemist~\cite{video_alchemist}, we report Masked DINO~\cite{dino} similarity (M-DINO) between each reference image and the masked subject in on-screen frames.
    \item \textit{Global caption following}: we report Masked CLIP~\cite{clip} similarity (M-CLIP) between each per-entity global description and the masked subject.
    \item \textit{Dense caption following}: we report ViCLIP~\cite{internvid,internvideo} similarity between each dense caption and the video frames inside its annotated interval. It is reported separately for four categories: subject (masked region), scene (entire frame), camera, and shot transition.
    \item \textit{Shot-transition timing}: we report a recall rate, where a Qwen2.5-VL-7B-Instruct~\cite{qwenvl2} judge evaluates whether each ground-truth transition occurs within a tolerance window.
\end{itemize}

\myparagraph{Quantitative comparison.}
\cref{tab:quantitative_real} reports the comparison on \ourbench, and shows a clear split between single-axis specialization and joint cinematic control. The strongest multi-subject personalization baselines remain competitive on subject identity and global appearance (\ie, Phantom~\cite{phantom} and VACE~\cite{vace} perform the best on M-DINO and M-CLIP, respetively), but these advantages do not carry over to dense caption following. The multi-shot baselines also do not dominate transition timing and remain weaker on identity preservation. In contrast, \ourmodel is strongest on the axes that require routing each condition to a specific entity and temporal interval, which is precisely the regime targeted by the unified entity-centric primitive.
\cref{app:additional_results} shows consistent trends on \ourbenchsyn.

\myparagraph{Qualitative comparison.}
The qualitative comparisons in \cref{fig:qualitative_main} mirrors the pattern in the quantitative study. Personalization baselines~\cite{phantom,vace} better preserve the identity but collapse the timeline into a single shot, while multi-shot baselines~\cite{cinetrans,echoshot,multishotmaster,shotstream} place cuts but drift on identity and dish continuity across them. \ourmodel is the only method that simultaneously preserves all four identities and hits three hard cuts. \cref{app:additional_results} provides more additional comparisons on \ourbenchsyn.

\begin{table}[t]
    \centering
    \mycaption{Ablation of two coordinated RoPEs on \ourbench}{
        From top to bottom, (a) applies AlcheMinT's 3-point WeRoPE~\cite{alchemint};
        (b) replaces WeRoPE with our $N{=}16$ interval-sampled temporal RoPE;
        (c) adds the entity-index axis to image refs and global caption tokens;
        (d) further restricts the entity axis on global caption to its visible-time intervals;
        (e) extends the entity axis to video tokens at visible entity rows;
        (f) promotes dense-event tokens into entity slots without duration-dependent rescaling.
        Best in \textbf{bold}.
    }
    \vspace{+2mm}
    \label{tab:ablation}
    \footnotesize
    \setlength{\tabcolsep}{1.7pt}
    \begin{tabular}{l|c|c|cccc|c}
        \toprule
        & \multicolumn{1}{c|}{\textbf{Subject}} & \multicolumn{1}{c|}{\textbf{Global Cap.}} & \multicolumn{4}{c|}{\textbf{Dense Cap. Following} (ViCLIP$\uparrow$)} & \textbf{Trans.} \\
        \cmidrule(lr){2-2}\cmidrule(lr){3-3}\cmidrule(lr){4-7}\cmidrule(lr){8-8}
        \textbf{Variant} & M-DINO$\uparrow$ & M-CLIP$\uparrow$ & Subject & Scene & Camera & Trans. & Recall$\uparrow$ \\
        \midrule
        \multicolumn{8}{l}{\textit{Interval-Sampled Temporal RoPE}} \\
        (a) WeRoPE~\cite{alchemint} ($1$ pos and $2$ neg), temp-axis on $\mathbf{V}$ & 0.455 & 0.285 & \textbf{0.236} & 0.206 & 0.192 & 0.145 & 0.399 \\
        (b) Ours interval-sampled ($N{=}16$), temp-axis on $\mathbf{V}$              & 0.477 & 0.289 & 0.234 & 0.208 & 0.191 & 0.147 & 0.406 \\
        \midrule
        \multicolumn{8}{l}{\textit{2D Entity-Temporal RoPE}} \\
        (c) +  entity-axis on $\mathbf{I}_k$ and $g_k$                               & 0.485 & 0.287 & 0.234 & 0.207 & 0.191 & 0.147 & 0.422 \\
        (d) ~~~~+ temp-axis on $\mathbf{I}_k$ and $g_k$                              & 0.484 & 0.288 & 0.233 & 0.206 & 0.190 & 0.148 & 0.394 \\
        (e) ~~~~~~~~+ entity-axis on $\mathbf{V}$                                    & 0.489 & 0.289 & 0.234 & 0.207 & 0.190 & 0.145 & 0.416 \\
        \midrule
        \multicolumn{8}{l}{\textit{Similarity-Peak Normalization}} \\
        (f) Full 2D entity-temp and no rescaling ($\beta(L)\!\equiv\!1$)             & 0.477 & 0.289 & 0.235 & 0.207 & 0.190 & 0.147 & 0.412 \\
        \midrule
        \rowcolor{gray!12}
        \textbf{\textit{Full method}}                                                & \textbf{0.502} & \textbf{0.295} & 0.235 & \textbf{0.208} & \textbf{0.193} & \textbf{0.150} & \textbf{0.486} \\
        \bottomrule
    \end{tabular}
\end{table}

\begin{figure*}[t]
    \centering
    \includegraphics[width=\linewidth]{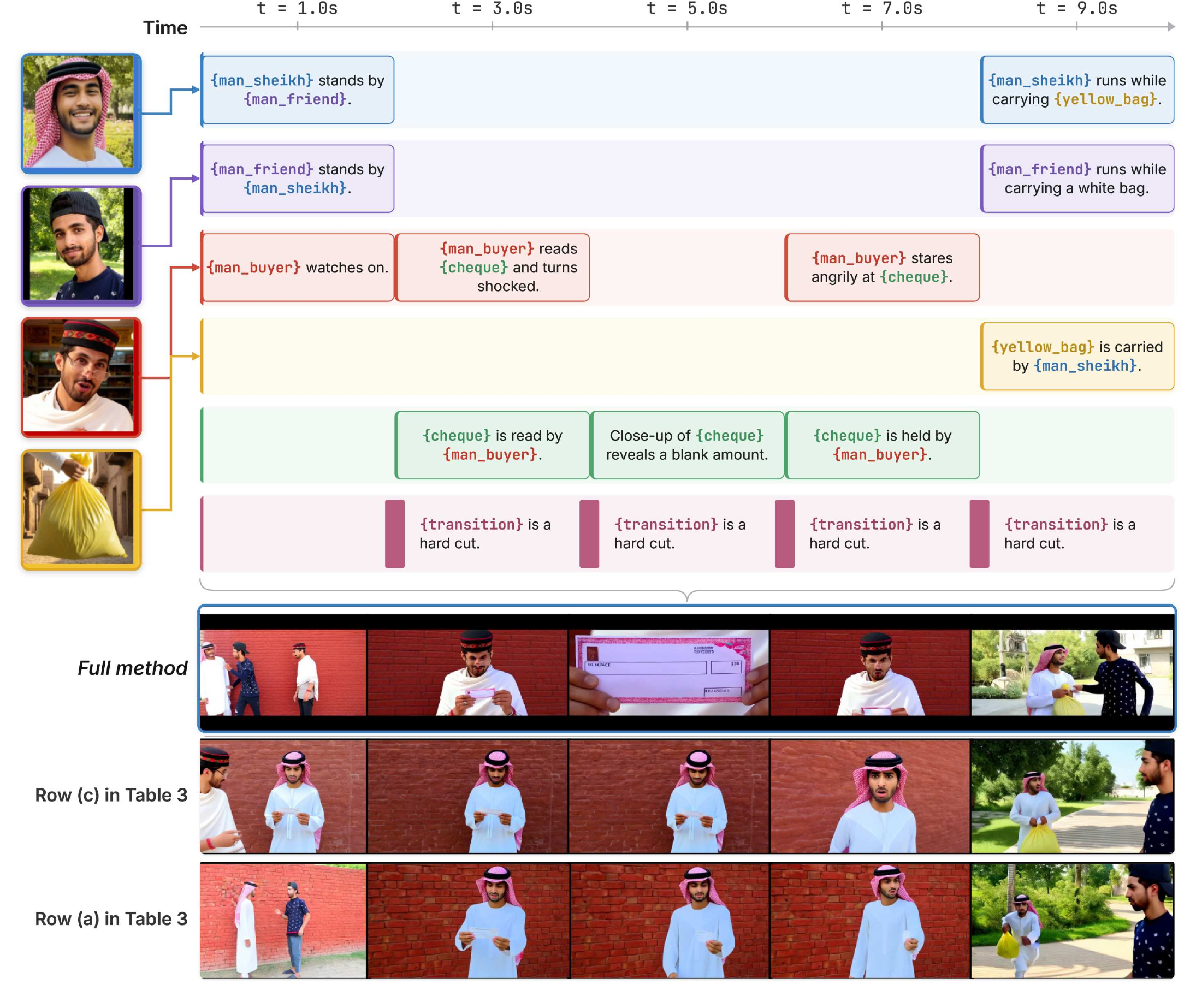}
    \mycaption{Visual ablation of two coordinated RoPEs}{
        Only the full method (top video row) routes each entity to its annotated interval and lands all four hard cuts at the specified times.
    }
    \label{fig:qualitative_ablation}
\end{figure*}

\myparagraph{User study.}
Since automatic metrics are coarse on perceptual properties, we complement them with a pairwise human evaluation on the same \ourbench, where raters score \ourmodel against each baseline on eight dimensions (see \cref{app:evaluation:user_study} for the full protocol). \cref{fig:user_study} shows that raters prefer \ourmodel on every entity-, text-, and structure-related dimension across all six baselines, and on perceptual dimensions (\textit{motion}, \textit{overall quality}, \textit{scene}) against most baselines.

\subsection{Ablation Study}
\label{sec:exp:ablation}

\cref{tab:ablation} and \cref{fig:qualitative_ablation} ablates the two core RoPE designs of \ourmodel on \ourbench. 
\begin{itemize}[leftmargin=*,itemsep=0pt,topsep=0pt,parsep=0pt,partopsep=0pt]
    \item From (a) to (b), replacing WeRoPE~\cite{alchemint} with interval-sampled temporal RoPE improves identity consistency and transition recall, while leaving dense-caption following largely stable.
    \item From (c) to (e), adding the entity-temporal RoPE gives additional gains, although the partial variants are not uniformly better across all metrics, suggesting that entity disambiguation alone is insufficient without the complete coordinate design.
    \item Comparing (f) with the full method, removing the $\beta(L)$ rescaling weakens identity consistency and transition timing, supporting the role of $\beta(L)$ for events with different temporal spans.
    \item \cref{fig:qualitative_ablation} further verifies that only the full method (top video row) routes every entity to its annotated interval and lands all four hard cuts at the prescribed timestamps.
\end{itemize}

\section{Conclusion}
\label{sec:conclusion}

We have presented \ourmodel, the first video diffusion framework to jointly control four cinematic axes within a single model: \textit{subjects}, \textit{events}, \textit{camera}, and \textit{shot transitions}.
At its core is an entity-centric conditioning primitive that represents every cinematic element as a unified, structured expression.
To realize this primitive inside a video DiT, we introduce two coordinated rotary embeddings that handle duration-varying events and route each per-entity condition to its target spatiotemporal region.
On \ourbench and \ourbenchsyn, \ourmodel outperforms six per-axis specialists on dense caption following and shot-transition timing, with consistent gains in a pairwise user study and ablations of the two coordinated RoPEs.
By unifying disparate cinematic controls, \ourmodel reconciles fine-grained controllability with long-horizon narrative coherence, opening the door to movie-level video generation directed by full script rather than single prompt.

\section{Limitations and Societal Impacts}
\label{app:limitations}

We identify the following limitations of our work:

\myparagraph{Coarse-grained cinematographic control. }
\ourmodel conditions camera motion and shot transitions through natural-language descriptors, which trade fine-grained geometric precision for accessibility. Applications demanding exact viewpoint repeatability or downstream 3D reconstruction are better served by trajectory-based methods~\cite{cameractrl,ac3d,cami2v,motionctrl} or source-video re-rendering~\cite{recammaster}. We leave combining our entity-centric prompts with explicit pose conditioning to future work.

\myparagraph{Bounded clip length.}
\ourmodel generates an entire scene in a single forward pass, which delivers strong cross-shot coherence but inherits the context-length limits of the underlying DiT. Multi-shot pipelines~\cite{cinetrans,echoshot,multishotmaster,shotstream} reach longer durations by synthesizing shots independently and stitching them post hoc, but suffer from weaker subject and lighting continuity. We see extending our primitive to long-form generation via autoregressive models with shared entity tokens as a next step.

\myparagraph{No audio modality.}
\ourmodel is purely visual, yet cinematic experience is inseparable from dialogue and ambient sound. Integrating an audio branch that respects the same entity-temporal conditioning is a promising future direction.

\myparagraph{Impact statement.}
\ourmodel lowers the barrier to producing cinematic video by letting users direct subjects, events, camera moves, and shot transitions through natural-language scripts and reference images. This can benefit independent filmmakers, educators, and accessibility-driven content creation. Faster previsualization may also reduce waste in physical productions.

At the same time, controllable identity-preserving video generation carries real risks of misuse shared with the broader class of generative video models, including non-consensual portrayal of real people, fabrication of harmful scenarios involving minors, and large-scale visual disinformation.

We recommend that any deployed system combine \textit{(i)} consent verification and watermarking, \textit{(ii)} reference-image policies blocking public figures and known minors, \textit{(iii)} prompt-level filters for sexual, violent, or politically deceptive content, and \textit{(iv)} gated weight release with abuse-investigation logging.

\clearpage
{
    \small
    \bibliographystyle{ieee_fullname}
    \bibliography{main}
}

\clearpage
\appendix

\begin{center}
    {\Large\bfseries \makeatletter\@title\makeatother\par}
    \vspace{3mm}
    {\Large Supplementary Material}
    \vspace{1em}
\end{center}

\section{Derivation and Properties of \texorpdfstring{$\beta(L)$}{β(L)}}
\label{app:beta}

\begin{figure}[t]
    \centering
    \includegraphics[width=\linewidth]{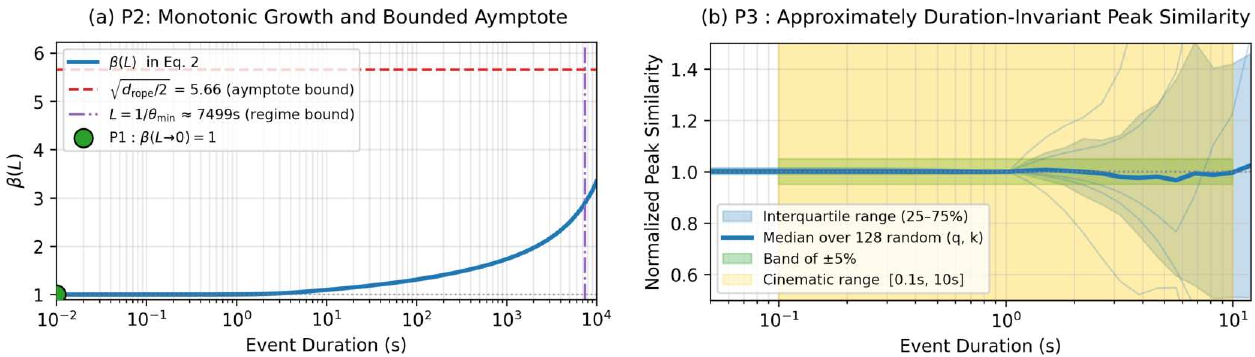}
    \mycaption{Closed-form $\beta(L)$ and its three properties}{
        (a) shows $\beta(L)$ from Eq.~\eqref{eq:beta_closed_form} across event durations $L$ in log-scale. The green marker verifies \textbf{P1} ($\beta(0)\!=\!1$). The curve is monotone (\textbf{P2}), upper-bounded by $\sqrt{d_{\text{rope}}/2}$ (red dashed) and approaches this asymptote only past $L\!\approx\!1/\theta_{\min}$ (purple dashed), so the cinematic range sits well within the bounded regime.
        (b) verifies \textbf{P3}: normalized peak similarity $\max_{t_v} s(t_v;e)$ (median and 25--75\% range over 128 random $(\mathbf{q},\mathbf{k})$ pairs) stays within $\pm 5\%$ (green band) across $[\text{0.1s}, \text{10s}]$ (yellow region).
    }
    \label{fig:beta}
\end{figure}

This appendix derives the closed form of the duration-dependent scaling $\beta(L)$ from \cref{eq:beta_closed_form} and proves its three properties: (P1) zero-duration limit $\beta(0) = 1$, (P2) monotonic growth toward a bounded asymptote, and (P3) approximate duration-invariance of the peak similarity $\max_{t_v} s(t_v; e)$. \cref{fig:beta} summarizes all three.

\myparagraph{Setup.}
Standard RoPE~\cite{rope} in $d_{\text{rope}}$ dimensions is the block-diagonal rotation
\begin{equation}
    R(t) \;=\; \mathrm{diag}\big(R_1(\theta_1 t),\, R_2(\theta_2 t),\, \dots,\, R_{d_{\text{rope}}/2}(\theta_{d_{\text{rope}}/2} t)\big),
    \label{eq:rope_block}
\end{equation}
where each $R_n(\alpha) \in \mathrm{SO}(2)$ is a planar rotation by angle $\alpha$ and $\{\theta_n\}_{n=1}^{d_{\text{rope}}/2}$ are the per-channel RoPE frequencies. Each block is orthogonal, so $\|R(t)\|_F^2 = d_{\text{rope}}$ for every $t$.
The dot product between a video token at time $t_v$ and an event token is
\begin{equation}
    s(t_v;\, e) \;=\; \frac{\beta(L)}{N}\sum_{i=0}^{N-1}\mathbf{q}^{\top} R(t_i - t_v)\,\mathbf{k},
    \label{eq:dot_product}
\end{equation}
where $\mathbf{q}, \mathbf{k} \in \mathbb{R}^{d_{\text{rope}}}$ are the query and key projections of the video and event-caption tokens.

\myparagraph{Closed form.}
Identifying each $\mathrm{SO}(2)$ block with the complex unit circle via $R_n(\alpha) \leftrightarrow e^{j\alpha}$, the $n$-th block of the unnormalized average $\tfrac{1}{N}\sum_{i=0}^{N-1} R(t_i)$ corresponds to
\begin{equation}
    z_n(L) \;=\; \frac{1}{N}\sum_{i=0}^{N-1} e^{j\theta_n t_i}.
\end{equation}
Writing $\bar t = (t_s + t_e)/2$, the offsets $t_i - \bar t$ are evenly spaced over $[-L/2,\, L/2]$, so the Dirichlet-kernel identity yields
\begin{equation}
    |z_n(L)| \;=\; \frac{1}{N}\,\Big|\frac{\sin\!\big(N \theta_n L / [2(N\!-\!1)]\big)}{\sin\!\big(\theta_n L / [2(N\!-\!1)]\big)}\Big|
    \;\xrightarrow[N\to\infty]{}\; |\mathrm{sinc}(\theta_n L / 2)|,
    \label{eq:dirichlet_block}
\end{equation}
with $\mathrm{sinc}(x) := \sin(x)/x$. We adopt the continuous-$N$ limit below; for $N=16$ and the RoPE frequencies considered here, the discrete and continuous expressions agree to within $1\%$.

The squared Frobenius norm of the unnormalized average is the sum of the per-block contributions $\|R_n\|_F^2 = 2$ scaled by $|z_n(L)|^2$:
\begin{equation}
    \Big\|\tfrac{1}{N}\sum_{i} R(t_i)\Big\|_F^2 \;=\; 2 \sum_{n=1}^{d_{\text{rope}}/2} \mathrm{sinc}^2(\theta_n L / 2).
    \label{eq:frob_closed}
\end{equation}
Imposing $\|\mathbf{P}^{\text{evt}}(e)\|_F^2 = \beta(L)^2 \cdot 2 \sum_n \mathrm{sinc}^2(\theta_n L/2) = d_{\text{rope}}$ yields the closed form
\begin{equation}
    \beta(L) \;=\; \Big(\tfrac{2}{d_{\text{rope}}}\textstyle\sum_{n=1}^{d_{\text{rope}}/2} \mathrm{sinc}^2(\theta_n L / 2)\Big)^{-1/2}.
    \label{eq:beta_app}
\end{equation}
\cref{eq:beta_app} depends only on $L$ and the RoPE spectrum $\{\theta_n\}$ and can be precomputed once.

\myparagraph{Property 1: zero-duration limit.}
At $L=0$, all sample positions coincide ($t_i = t_s$ for every $i$), so $\mathrm{sinc}(0) = 1$ and the sum in \cref{eq:beta_app} equals $d_{\text{rope}}/2$. Substituting gives $\beta(0) = (\tfrac{2}{d_{\text{rope}}} \cdot \tfrac{d_{\text{rope}}}{2})^{-1/2} = 1$, and $\mathbf{P}^{\text{evt}}(e)$ collapses to the single rotation $R(t_s)$, recovering standard single-position RoPE (green marker in \cref{fig:beta}(a)).

\myparagraph{Property 2: monotonic growth and bounded asymptote.}
Each $\mathrm{sinc}^2(\theta_n L/2)$ is non-increasing on $[0,\, 2\pi/\theta_n]$, the regime in which the corresponding frequency component has not yet completed its first arc. Across this range $\sum_n \mathrm{sinc}^2(\theta_n L/2)$ is non-increasing, so $\beta(L)$ is non-decreasing in $L$. As $L$ grows past this range, high-frequency components ($\theta_n L \gtrsim 2\pi$) average out and only the slowest frequencies contribute meaningfully to the sum. In the practical regime $\theta_{\min} L \ll 1$ (satisfied for all event durations encountered in this paper given standard RoPE frequencies), $\beta(L)$ is therefore upper-bounded by
\begin{equation}
    \beta(L) \;\le\; \sqrt{d_{\text{rope}}/2},
    \label{eq:beta_bound}
\end{equation}
attained when only the smallest RoPE frequency contributes the full $|\mathrm{sinc}(\theta_{\min} L/2)| \approx 1$. \cref{fig:beta}(a) plots $\beta(L)$ (blue) against this asymptote (red dashed) and the regime bound $L \approx 1/\theta_{\min}$ (purple dashed), showing the cinematic range sits well inside the bounded regime.

\myparagraph{Property 3: approximate duration-invariance of the peak similarity.}
Decomposing $\mathbf{q}, \mathbf{k}$ into per-block 2D components $\mathbf{q}_n, \mathbf{k}_n \in \mathbb{R}^2$ and writing $\mathbf{q}_n^{\top} R_n(\alpha)\mathbf{k}_n = a_n \cos(\alpha + \phi_n)$ for amplitudes $a_n = \|\mathbf{q}_n\|\,\|\mathbf{k}_n\|$ and phases $\phi_n$, the similarity profile of \cref{eq:dot_product} expands as
\begin{equation}
    s(t_v;\, e) \;=\; \beta(L) \sum_{n=1}^{d_{\text{rope}}/2} a_n\, \mathrm{sinc}\!\big(\theta_n L / 2\big)\, \cos\!\big(\theta_n[\bar t - t_v] + \phi_n\big).
    \label{eq:sim_app}
\end{equation}
At $t_v = \bar t$ the cosine factor reduces to $\cos(\phi_n)$ for every $n$, and bounds on $\max_{t_v} s(t_v; e)$ follow from how the per-frequency amplitudes $\{a_n\}$ are distributed across the spectrum.

\textit{(i) Spectrally diffuse $\mathbf{q}, \mathbf{k}$.}
When the energy $\{a_n^2\}$ is spread over multiple frequencies, the Cauchy--Schwarz inequality gives
\begin{equation}
    \big| s(\bar t;\, e) \big| \;\le\; \beta(L) \sqrt{\textstyle\sum_n a_n^2}\, \sqrt{\textstyle\sum_n \mathrm{sinc}^2(\theta_n L / 2)}.
    \label{eq:cs_bound}
\end{equation}
Substituting \cref{eq:beta_app} cancels the second square-root factor exactly, leaving
\begin{equation}
    \big|s(\bar t;\, e)\big| \;\le\; \sqrt{d_{\text{rope}}/2}\, \sqrt{\textstyle\sum_n a_n^2},
\end{equation}
which is independent of $L$. The peak similarity is therefore \emph{exactly} $L$-invariant in the spectrally diffuse regime.

\textit{(ii) Spectrally concentrated $\mathbf{q}, \mathbf{k}$.}
When most of the energy concentrates on a single frequency $\theta_{n^\star}$,
\begin{equation}
    \big|s(\bar t;\, e)\big| \;\approx\; a_{n^\star}\, \beta(L)\, \big|\mathrm{sinc}(\theta_{n^\star} L / 2)\big|,
\end{equation}
and invariance holds up to a multiplicative factor that depends on whether $\theta_{n^\star}$ lies in the slow- or fast-decaying part of the spectrum. The factor is bounded above by $\sqrt{d_{\text{rope}}/2}$ in either case (Property~2).

Learned attention patterns sit between regimes \textit{(i)} and \textit{(ii)} in practice. \cref{fig:beta}(b) confirms empirically that $\max_{t_v} s(t_v; e)$ stays within $\pm 5\%$ across the full cinematic range $L \in [\text{0.1s}, \text{10s}]$.

\section{Implementation Details}
\label{app:implementation}

\myparagraph{Data samples.}
\label{app:implementation:dataset}
Our training corpus consists of videos ranging from one minute to two hours with wide-varying camera angles/shots/multiple entities across multiple scenes. Each title is split into non-overlapping one-minute chunks that serve as atomic training samples. We run a YOLO~v11 person detector~\cite{yolo} on one frame every $10$~s and discard chunks whose majority of probe frames have either zero people or a count above a fixed clutter threshold. A $2\%$ random sample of training chunks (with overlapping source titles) is held out only as a loss-monitoring validation set. Our test benchmark \ourbench is built from a disjoint pool of source titles with no character or location overlap with training; we manually curate $512$ one-minute chunks from this pool.

\subsection{Annotation of Entity-Centric Cinematic Conditioning}
\label{app:implementation:annotation}
Each chunk is annotated through entity-centric captioning followed by reference-image curation with appearance augmentation.

\begin{figure*}[t]
    \centering
    \includegraphics[width=\linewidth]{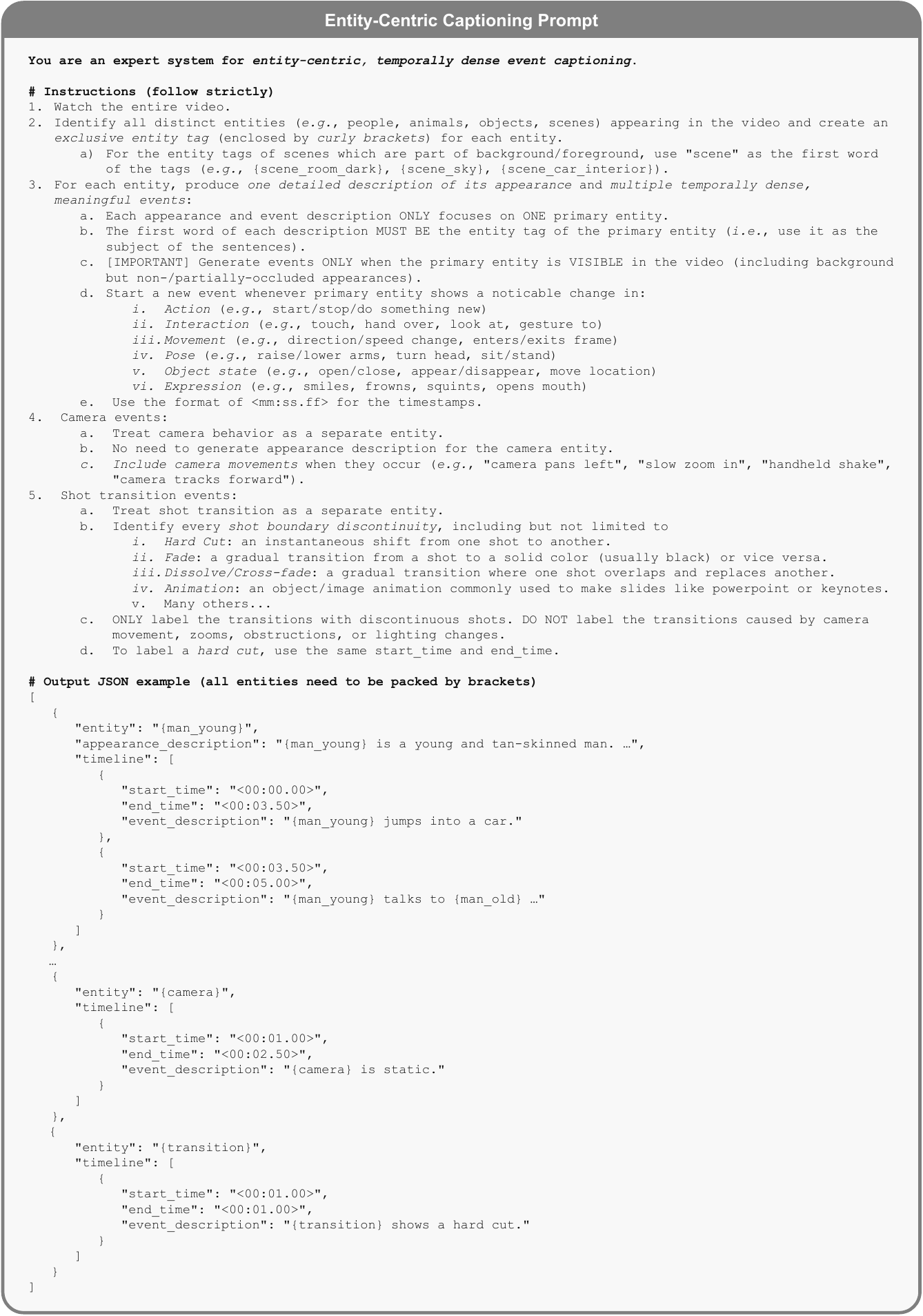}
    \mycaption{Entity-centric captioning prompt}{
        A single structured-output query to Gemini-2.5-Pro~\cite{gemini} returns entity tags, per-entity global descriptions, and dense \texttt{<mm:ss.ff>}-stamped event timelines for visual subjects, \texttt{\{camera\}}, and \texttt{\{transition\}} in one pass.
    }
    \label{fig:gemini_prompt}
\end{figure*}

\myparagraph{Entity-centric captioning.}
A single Gemini-2.5-Pro~\cite{gemini} call (prompt in \cref{fig:gemini_prompt}, fixed JSON schema) returns a unique curly-brace tag per distinct entity (\eg, \texttt{\{girl\_young\}}, \texttt{\{scene\_museum\}}), a one-sentence global appearance description per visual subject, and a temporally dense per-entity timeline of \texttt{<mm:ss.ff>}-stamped events whose descriptions begin with the primary entity tag. Two cinematography entities are tracked alongside the visual subjects: \texttt{\{camera\}} (pans, zooms, tracks, handheld) and \texttt{\{transition\}} (hard cuts, fades, dissolves, animated wipes).

\myparagraph{Reference-image curation with appearance augmentation.}
For each visual subject we sample one frame at the centre of each event and prompt Gemini-3-Flash-Preview~\cite{gemini} for a bounding box, reusing the global description as a disambiguating clue (entities with no returned box are dropped). We crop each box, encode it with CLIP ViT-L/14~\cite{clip}, and rank crops by cosine similarity against the entity's global description; we keep the top four and sample two without replacement with probability proportional to score. The pair is passed jointly into Qwen-Image-Edit~\cite{qwen_image_edit} with an entity-aware prompt that places the subject in a new background, lighting, and pose (also varying expression for people-like entities). The resulting image is the appearance-augmented reference $\mathbf{I}_k$ used during training;

\myparagraph{Per-chunk caption statistics of \ourbench.}
Each chunk averages $4.6$ visual subject entities and $13.4$ events ($8.7$ subject, $2.7$ camera, $2.0$ transitions). Event durations are heavily right-skewed (median $2.5$~s, $10^\text{th}$~pct.\ $0.7$~s, $90^\text{th}$~pct.\ $10.2$~s) with a long tail of $L\!=\!0.1$ shot-transition events. This combination of many entities, many events per entity, and per-event durations spanning four orders of magnitude motivates the duration-aware $\beta(L)$ rescaling in \cref{sec:method:event_rope}.

\subsection{Model Architecture}
\label{app:implementation:architecture}
We use the Wan~2.1 video autoencoder~\cite{wan} (frozen, $8{\times}8$ spatial and $4{\times}$ temporal compression, $16$ latent channels) to encode video, and apply it identically to reference images as $1$-frame clips. The backbone is a pre-LN diffusion transformer~\cite{dit} that operates on the video latent concatenated with reference-image tokens; each block applies self-attention over the joint sequence, cross-attention to the text-only streams, and a gated MLP, with time-step and scalar conditioning projected to per-block modulation weights. All captions are encoded with T5-XXL~\cite{t5} ($4096$-dim/token); per chunk we encode up to $16$ entities (with $128$-token global descriptions) and up to $32$ dense events (with $64$-token event descriptions). The two streams stay separate so the entity-temporal RoPE (\cref{sec:method:event_rope}) can address them under different coordinate conventions, and empty slots are zero-padded with attention masking. Each chunk additionally carries up to $K\!=\!4$ appearance-augmented references $\mathbf{I}_k$ (\cref{app:implementation:annotation}), one per dominant entity; unused slots are zero-filled and masked.

\clearpage

\clearpage

\subsection{Training and Inference}
\label{app:implementation:training}
\myparagraph{Pretraining and fine-tuning.}
We first pretrain the backbone on a large corpus of generic video--caption pairs at the same architecture and latent space, with the entity-temporal RoPE and reference-image inputs disabled, then fine-tune \ourmodel from this checkpoint on the entity-annotated chunks of \cref{app:implementation:dataset}. All hyperparameters below refer to the fine-tuning stage. Each step draws a $10$-second window from a one-minute chunk and decodes it at the native $15$~fps, yielding $153$-frame clips at $288{\times}512$.

\myparagraph{Optimizer, schedule, and objective.}
We use AdamW~\cite{adamw} ($\beta_1\!=\!0.9$, $\beta_2\!=\!0.99$, $\epsilon\!=\!10^{-8}$, weight decay $0.01$); most parameters train at $3\!\times\!10^{-5}$, while the cross-attention layers reorganised under the 2D entity-temporal RoPE (\cref{sec:method:cross_attn}) train at $1\!\times\!10^{-4}$. The schedule is constant after a $1{,}000$-step linear warmup with global gradient-norm clip $1.0$. Mixed precision uses \texttt{bfloat16} parameters and \texttt{float32} gradient reductions, sharded via FSDP-2~\cite{fsdp} with per-block activation checkpointing; an EMA copy at $\beta_{\mathrm{ema}}\!=\!0.9999$ (after $1{,}000$-step warmup) is used for all reported numbers and samples. We train rectified flow~\cite{flow_matching, rectified_flow} with $\sigma_{\mathrm{data}}\!=\!1$, $\sigma_{\mathrm{noise}}\!=\!2$, logit-normal timesteps (location $0$, scale $1$, $\epsilon\!=\!10^{-3}$, logit-shift $3.0$) and the EDMv2-normalised regression~\cite{edm2} with image/video weights both $1.0$. To enable separate inference-time guidance scales, we apply Bernoulli CFG-dropout~\cite{cfg} with probability $0.1$ to each of the reference-image stream, the joint global-and-dense caption stream, and the auxiliary scalar conditions (resolution, dataset id, sampling frame rate); the two text streams are dropped synchronously. Training lasts $25{,}000$ iterations on $32$ NVIDIA H100 GPUs at per-GPU batch size $2$.

\myparagraph{Multi-condition CFG and inference.}
\ourmodel takes two grouped conditioning modalities: text $c_{txt}$ (covering $g_k$ and $e_{k,j}$) and image $c_{img}$ (covering $\mathbf{I}_k$). We combine them as
\begin{align*}
\tilde{f}_{\theta}(z_t,c_{img},c_{txt})
&= \quad f_{\theta}(z_t,c_{img},c_{txt}) \\
&+ \lambda_{txt} \cdot \big(f_{\theta}(z_t,c_{img},c_{txt}) - f_{\theta}(z_t,c_{img},\varnothing)\big) \\
&+ \lambda_{img} \cdot \big(f_{\theta}(z_t,c_{img},c_{txt}) - f_{\theta}(z_t,\varnothing,c_{txt})\big) \\
&+ \lambda_{joint} \cdot \big(f_{\theta}(z_t,c_{img},c_{txt}) - f_{\theta}(z_t,\varnothing,\varnothing)\big),
\end{align*}
with $\varnothing$ the fully-unconditional pass and $\lambda_{joint}\!=\!5$, $\lambda_{img}\!=\!3$, $\lambda_{txt}\!=\!0$. We sample with a $40$-step rectified-flow sampler at the same $288{\times}512$ / $153$-frame ($10$~s at $15$~fps) shape used during training; in practice we can also generate $40$~s videos at $288{\times}512$ or $10$~s videos at $720{\times}1280$ as we show in our supplementary material and in~\cref{app:additional_results}. Timesteps are warped with a time-shifting factor of $5.66$ to spend more steps in the higher-noise region.

\section{Evaluation Details}
\label{app:evaluation}

\subsection{Curation of \ourbenchsyn}
\label{app:evaluation:benchsyn}

\ourbenchsyn complements the real-footage benchmark with $512$ hand-authored 10.2s scenarios. Each scenario emits the same JSON schema as the real-footage annotations (\cref{app:implementation:annotation}) --- \verb|global_entities| covering \verb|{camera}|, \verb|{transition}|, \verb|{scene_*}|, characters, animals, and props with their visible-time intervals, and \verb|dense_entities| listing fine-grained timestamped events --- so the same downstream pipeline serves both data sources. Authoring is parameterised across shot count, character count, and event density; zero-width hard cuts and the $\le\!16$ entity / $\le\!64$ event caps match training conventions, and a small validator enforces structural invariants (canonical scene names, no overlapping scenes, all referenced names resolve, intervals in $[0, 10.2]$). For every non-camera, non-transition, non-scene entity we generate a single reference image with the publicly released Qwen-Image text-to-image model~\cite{qwen_image_edit} at $1024{\times}1024$, $50$ inference steps, true-CFG-scale $4.0$, fixed seed, and a quality-oriented negative prompt. The image-generation prompt is built from the entity's appearance description by stripping curly braces, replacing underscores with spaces, choosing the relative pronoun \emph{who} or \emph{which} from a hand-curated person/animal token list, reading singular/plural number off the description's leading copula, and emitting \textit{``An image of \{a/an\}\,$\langle$natural name$\rangle$\,\{who$|$which\}\,\{is$|$are\}\,$\langle$description$\rangle$''}. The PNG path is written back into the JSON's \verb|ref_image_path| field so training, evaluation, and user-study pipelines locate it identically to a real-footage reference. Across the evaluated slice, scenarios average $4.5$ entities, $1.6$ events per entity, $3.1$ camera events, $2.1$ transitions, and $3.5$ reference images (max $6$), with per-event durations spanning 0.1s to 10.2s.

\subsection{Baselines}
\label{app:evaluation:baselines}

Each \ourmodel caption carries persistent entity descriptions, per-entity time intervals, per-entity reference images, and dense per-shot captions; adapting them to each baseline relies on four mechanisms. \emph{Shot boundaries} are derived from transition events ($N$-shot samples have $N{-}1$ transitions): hard cuts collapse to a single timestamp, while long fades are split at their midpoint and allocated half to each adjacent shot. \emph{Annotation trimming} aligns the rendered duration: events past the rendered window are dropped, span-crossing events are clipped, and orphaned entities are removed. ShotStream~\cite{shotstream}, MultiShotMaster~\cite{multishotmaster}, VACE~\cite{vace}, and Phantom~\cite{phantom} render the full sample and need no trimming; CineTrans~\cite{cinetrans} (5.06\,s) and EchoShot~\cite{echoshot} (7.81s) are trimmed to their fixed durations before prompt construction. \emph{Active-entity filtering} avoids T5's~\cite{t5} 512-token cap (silently truncated by Wan-family pipelines~\cite{wan}): the per-shot context for MultiShotMaster, EchoShot, VACE, and Phantom keeps only entities whose intervals overlap the current shot, while ShotStream and CineTrans retain their released schema (a flat sample-level global of brace-stripped concatenated entity descriptions). \emph{Reference-image captioning} is needed for the four text-only baselines (ShotStream, MultiShotMaster, CineTrans, EchoShot --- MultiShotMaster had only released text-to-video weights at writing): we generate one- to two-sentence captions of each reference image with Qwen3.5-35B-A3B~\cite{qwen3p5} (chain-of-thought disabled) and inline them as the entity description; VACE and Phantom consume reference images directly.

All baselines run with their public weights at $832{\times}480$ resolution and $16$~fps (MultiShotMaster $15$~fps), $50$ denoising steps (ShotStream $4$, using its distilled checkpoint). VACE and Phantom only support single-shot generation, so we run inference per shot and concatenate; CineTrans and EchoShot are restricted to fixed 5.06s and 7.81s outputs, while the rest match the per-shot lengths in our annotations.

\subsection{Evaluation Metrics}
\label{app:evaluation:metrics}

We evaluate every generated video on \ourbench (and \ourbenchsyn) against its conditioning annotations (entity tags, appearance descriptions, dense event timelines, transitions); the pipeline produces five metrics averaged across the $512$ videos --- two for entity identity, and one each for global-caption alignment, dense-caption alignment, and transition timing.

\myparagraph{Per-entity mask extraction.}
For every non-camera, non-transition entity we merge visible-time intervals into disjoint segments and sample three keyframes per segment (at the $20$/$50$/$80$\textsuperscript{th} percentiles). We run Grounding-DINO (\texttt{grounding-dino-base})~\cite{groundingdino} on each keyframe with the entity's natural-language tag (box threshold $0.25$, text threshold $0.20$), re-score the boxes with CLIP ViT-B/32~\cite{clip}, drop boxes with CLIP-text score below $0.1$, and feed the survivors to SAM2 (\texttt{sam2-hiera-large})~\cite{sam2} which produces both per-keyframe masks and a forward/backward-tracked mask video. The same crops, masks, and per-entity mask videos are reused by every downstream metric.

\myparagraph{Subject identity consistency (DINO, M-DINO).}
We encode each entity's reference image $\mathbf{I}_k$ with DINOv2~\cite{dinov2} and compute cosine similarity to every kept crop, taking the max per disjoint interval, the mean per entity, and the mean per video and benchmark. DINO uses raw bounding-box crops; M-DINO replaces the box background with black using the SAM2 keyframe mask. We report M-DINO as the headline subject-identity number.

\myparagraph{Global caption following (CLIP, M-CLIP).}
The same crops are encoded with CLIP ViT-B/32 and scored against the CLIP text embedding of the entity tag in natural language (e.g.\ \verb|{man_young}|\,$\to$\,\emph{``man young.''}). The unmasked / SAM-masked variants follow the same max-over-crops, mean-over-intervals, mean-over-entities aggregation; M-CLIP-text is reported.

\myparagraph{Dense caption following (ViCLIP).}
For every dense event we extract the corresponding sub-clip, sample eight uniformly-spaced frames, and compute cosine similarity between the ViCLIP-B/16 video embedding~\cite{internvid,internvideo} and the ViCLIP text embedding of the event description (with entity tags expanded to natural language). Events are bucketed into \emph{subject}, \emph{scene}, \emph{camera}, and \emph{transition} categories, and the mean-over-events-within-video, mean-over-videos statistic is reported per category to form the dense-caption block of \cref{tab:ablation,tab:quantitative_real}.

\myparagraph{Shot-transition timing (VLM recall).}
For each transition event we cut a sub-clip spanning $[\textit{start}-\delta_l,\, \textit{end}+\delta_r]$, with each side padded outward by up to $1.0$~s and clipped to maintain at least $0.5$~s of separation from the next adjacent transition (or clip boundary). The sub-clip is passed to Qwen2.5-VL-7B-Instruct~\cite{qwenvl2} with the event description and a yes/no prompt asking \textit{(i)} whether any shot transition is visible (\textsc{Present}) and \textit{(ii)} whether the type/style matches the description (\textsc{Matches}). The reported number is the per-video \emph{presence recall} averaged across the benchmark; the stricter \emph{match recall} is computed as a sanity check.

\subsection{User Study Protocol}
\label{app:evaluation:user_study}

Automatic metrics are coarse, especially for motion plausibility and aesthetic quality where neither CLIP- nor VLM-based evaluators are well calibrated. We therefore complement \cref{app:evaluation:metrics} with a head-to-head human evaluation on \ourbench; results are visualised in \cref{fig:user_study} and tabulated in \cref{tab:user_study}.

\myparagraph{Setup and rating.}
For every prompt, raters see two anonymised side-by-side videos (\textsc{Left}/\textsc{Right}, side randomised per item) generated from the same conditioning bundle: per-entity reference images, global appearance descriptions, the dense event timeline, and the special \verb|{camera}| / \verb|{transition}| tracks. They judge against the prompt alone (no ground-truth shown) on the five-point scale (\textit{1=strongly Left, 3=tie, 5=strongly Right}); ratings are required for every dimension and aggregated by averaging per dimension, with a method's win-rate reported as the fraction of items whose average favours that side.

\myparagraph{Evaluation dimensions.}
Eight dimensions are scored independently:
\begin{itemize}[leftmargin=*,itemsep=0pt,topsep=0pt,parsep=0pt,partopsep=0pt]
    \item \textit{Subject/entity consistency} judges face, body, and core object identity for entities with a reference image (clothing/accessory drift is not penalised; items with no reference images default to $3$).
    \item \textit{Global-description following} scores per-entity appearance attributes (clothing, hairstyle, accessories, body type) excluding face identity and \verb|{scene_*}| entities. 
    \item \textit{Dense-caption following} judges whether each event happens, with the right subject and roughly the right timing (excluding \verb|{camera}| and \verb|{transition}| events). 
    \item \textit{Shot structure} scores shot count, transition position and \emph{type} (a hard cut for a requested dissolve is penalised), and absence of spurious cuts.
    \item \textit{Camera angles and movement} scores pan/zoom direction, static-vs-moving, and continuous-move-vs-cut against the \verb|{camera}| timeline.
    \item \textit{Scene-description following} scores \verb|{scene_*}| coverage and timing. 
    \item \textit{Overall motion quality} judges smoothness and physical plausibility independent of caption following.
    \item \textit{Overall video quality} judges sharpness, composition, lighting, aesthetics, and frame-level cleanness.
\end{itemize}

\begin{figure}[t]
    \centering
    \includegraphics[width=\linewidth]{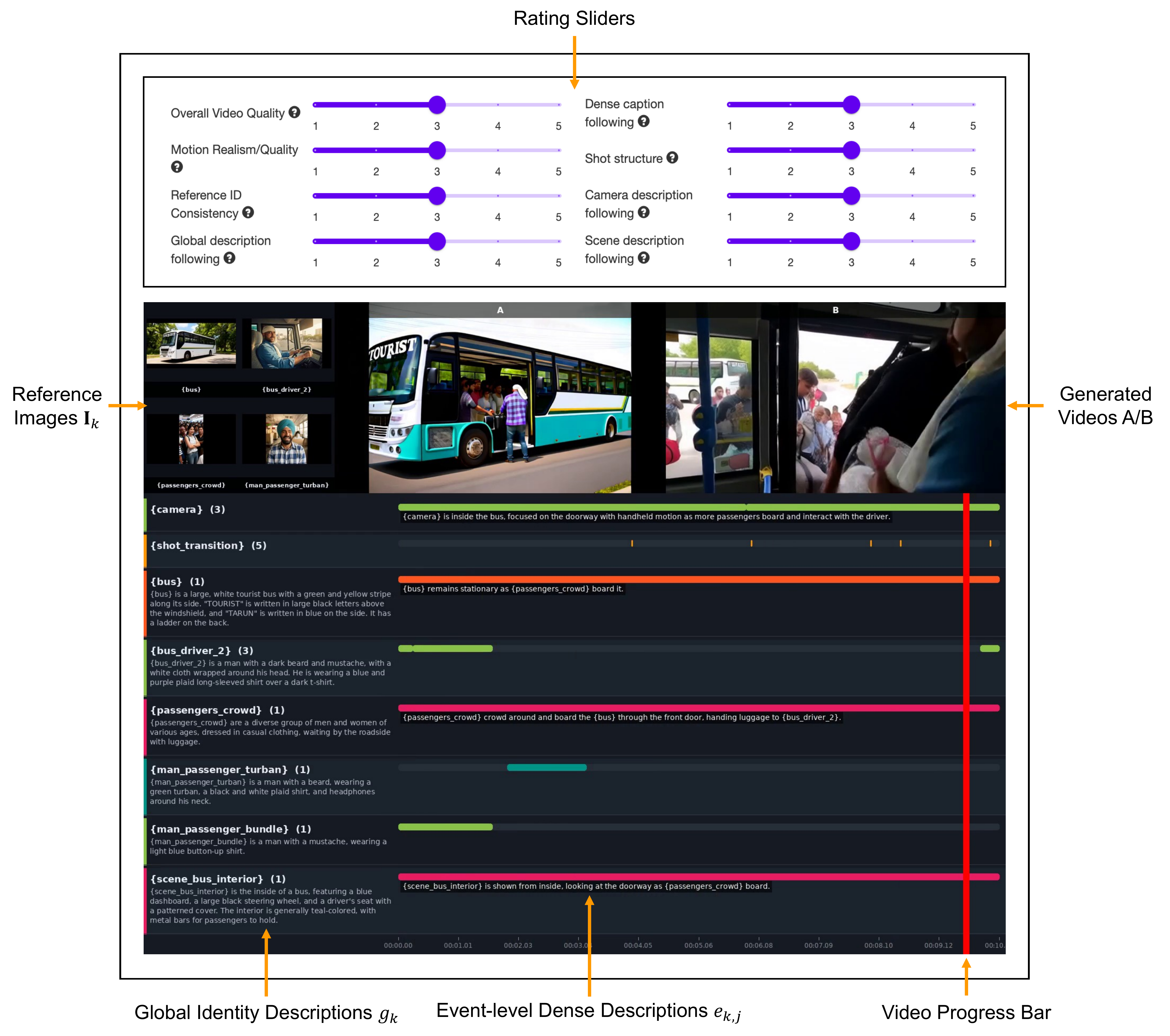}
    \mycaption{User study interface}{
        Side-by-side rating UI shown to in-house raters. Each video is scored on a $1-5$ scale across eight axes spanning visual quality (overall quality, motion realism), identity preservation (reference ID consistency), and prompt adherence (global description, dense caption, shot structure, camera, and scene). \cref{app:evaluation:user_study} shows full instructions and per-axis question.
    }
    \label{fig:user_study_UI}
\end{figure}

\myparagraph{Annotation interface.}
The task uses a video-annotation tool (\cref{fig:user_study_UI}) where dense events appear as time-track captions and global descriptions in a side panel; we extend the same labelling interface used to collect the underlying entity-centric annotations, so labellers see captions in the same layout they are rating against. Side assignment is randomised per item and method identity is hidden from raters.

\clearpage
\begin{table}[t]
    \centering
    \mycaption{Pairwise human-evaluation preferences on \ourbench (\%)}{
        Each cell reports raters' wins / ties / losses rate for \ourmodel against the indicated baseline along the corresponding dimension. \underline{Underline} marks dimensions where wins exceed losses.
        We report the original numbers in \cref{tab:user_study}
    }
    \vspace{+2mm}
    \label{tab:user_study}
    \footnotesize
    \setlength{\tabcolsep}{2.5pt}
    \begin{tabular}{l|cccccccc}
        \toprule
        \textbf{Method}
          & \textbf{Subject ID} & \textbf{Global Cap.} & \textbf{Dense Cap.}
          & \textbf{Camera}     & \textbf{Scene}       & \textbf{Shot Structure}
          & \textbf{Motion}     & \textbf{Overall} \\
        \midrule
        Phantom~\cite{phantom}                 & \underline{62/27/11} & \underline{54/37/9}  & \underline{82/13/5}  & \underline{70/26/4}  & \underline{44/35/21} & \underline{55/29/16} & \underline{79/16/5}  & \underline{79/14/7}  \\
        VACE~\cite{vace}                       & \underline{81/9/9}   & \underline{65/21/14} & \underline{73/14/14} & \underline{68/26/6}  & 28/33/38             & \underline{37/42/21} & \underline{65/20/15} & \underline{66/13/21} \\
        \midrule
        CineTrans~\cite{cinetrans}             & \underline{80/16/4}  & \underline{68/20/12} & \underline{75/16/9}  & \underline{54/39/7}  & 30/38/32             & \underline{26/59/15} & \underline{52/29/19} & \underline{46/21/33} \\
        EchoShot~\cite{echoshot}               & \underline{84/10/6}  & \underline{78/16/6}  & \underline{86/10/4}  & \underline{68/28/4}  & \underline{48/31/21} & \underline{40/44/16} & \underline{68/22/10} & \underline{65/18/17} \\
        MultiShotMaster~\cite{multishotmaster} & \underline{76/16/8}  & \underline{68/22/10} & \underline{75/16/9}  & \underline{55/34/10} & \underline{41/39/21} & \underline{36/42/21} & \underline{45/27/28} & \underline{44/22/34} \\
        ShotStream~\cite{shotstream}           & \underline{80/13/7}  & \underline{60/23/18} & \underline{74/19/7}  & \underline{58/34/8}  & 25/34/42             & \underline{49/32/19} & 29/33/38             & 32/14/54             \\
        \bottomrule
    \end{tabular}
\end{table}

\clearpage

\section{Additional Results}
\label{app:additional_results}

\begin{figure*}[t]
    \centering
    \includegraphics[width=\textwidth]{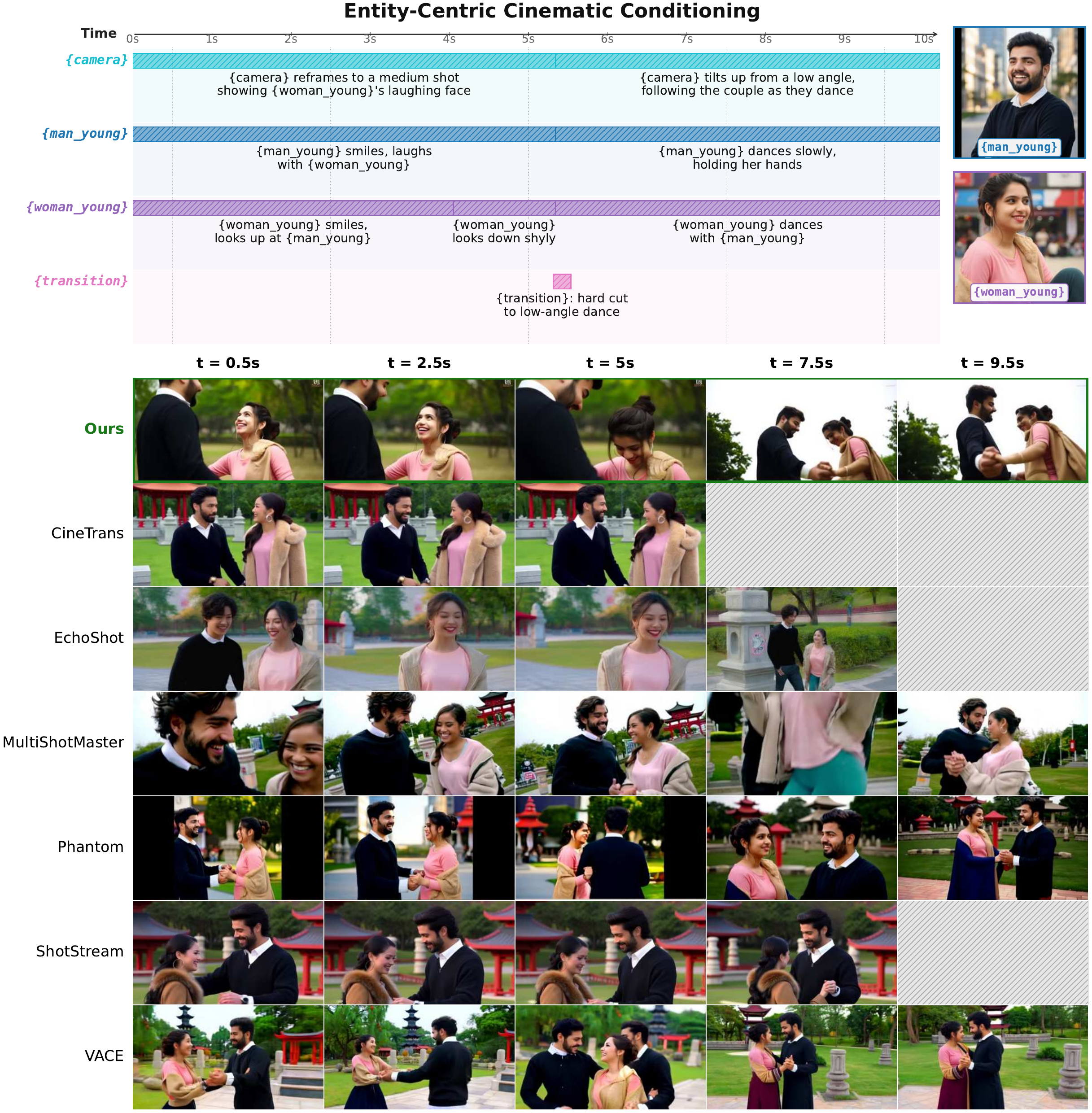}
    \mycaption{Additional qualitative comparison on \ourbench}{
    }
    \label{fig:qualitative_extra_real_1}
\end{figure*}

\begin{figure*}[t]
    \centering
    \includegraphics[width=\textwidth]{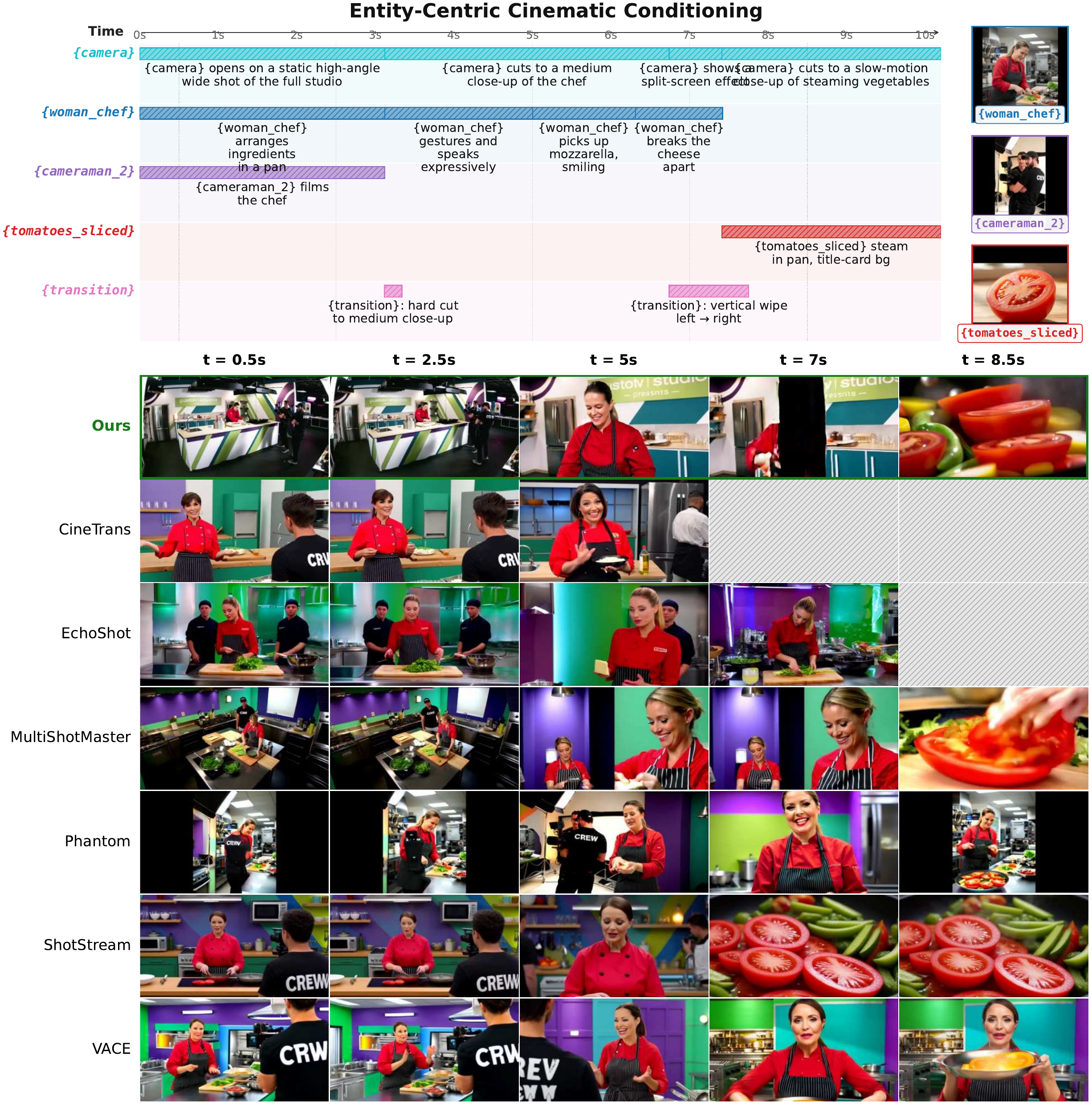}
    \mycaption{Additional qualitative comparison on \ourbench}{
    }
    \label{fig:qualitative_extra_real_2}
\end{figure*}

\begin{figure*}[t]
    \centering
    \includegraphics[width=\textwidth]{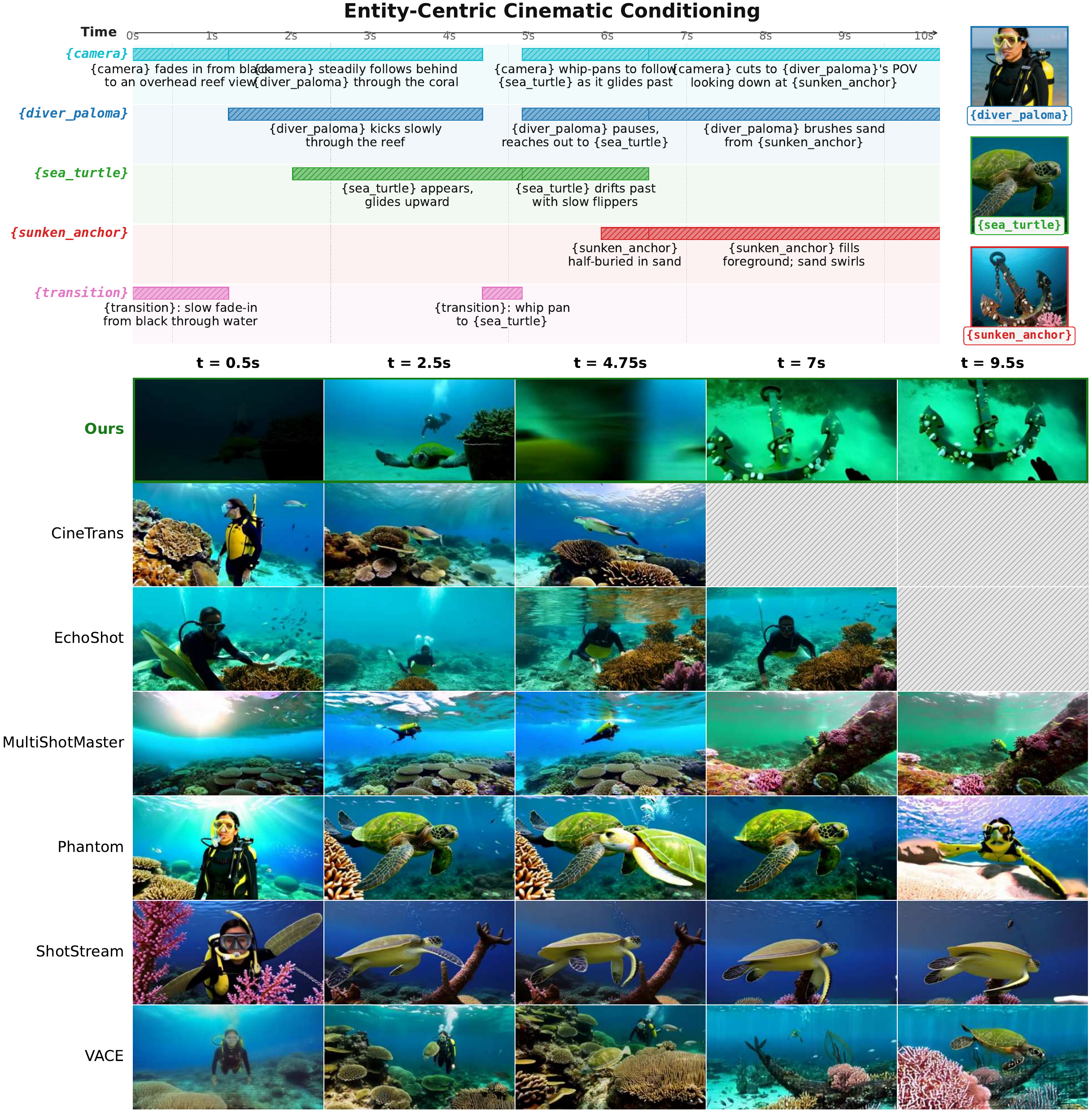}
    \mycaption{Additional qualitative comparison on \ourbenchsyn}{
    }
    \label{fig:qualitative_extra_syn_1}
\end{figure*}

\begin{figure*}[t]
    \centering
    \includegraphics[width=\textwidth]{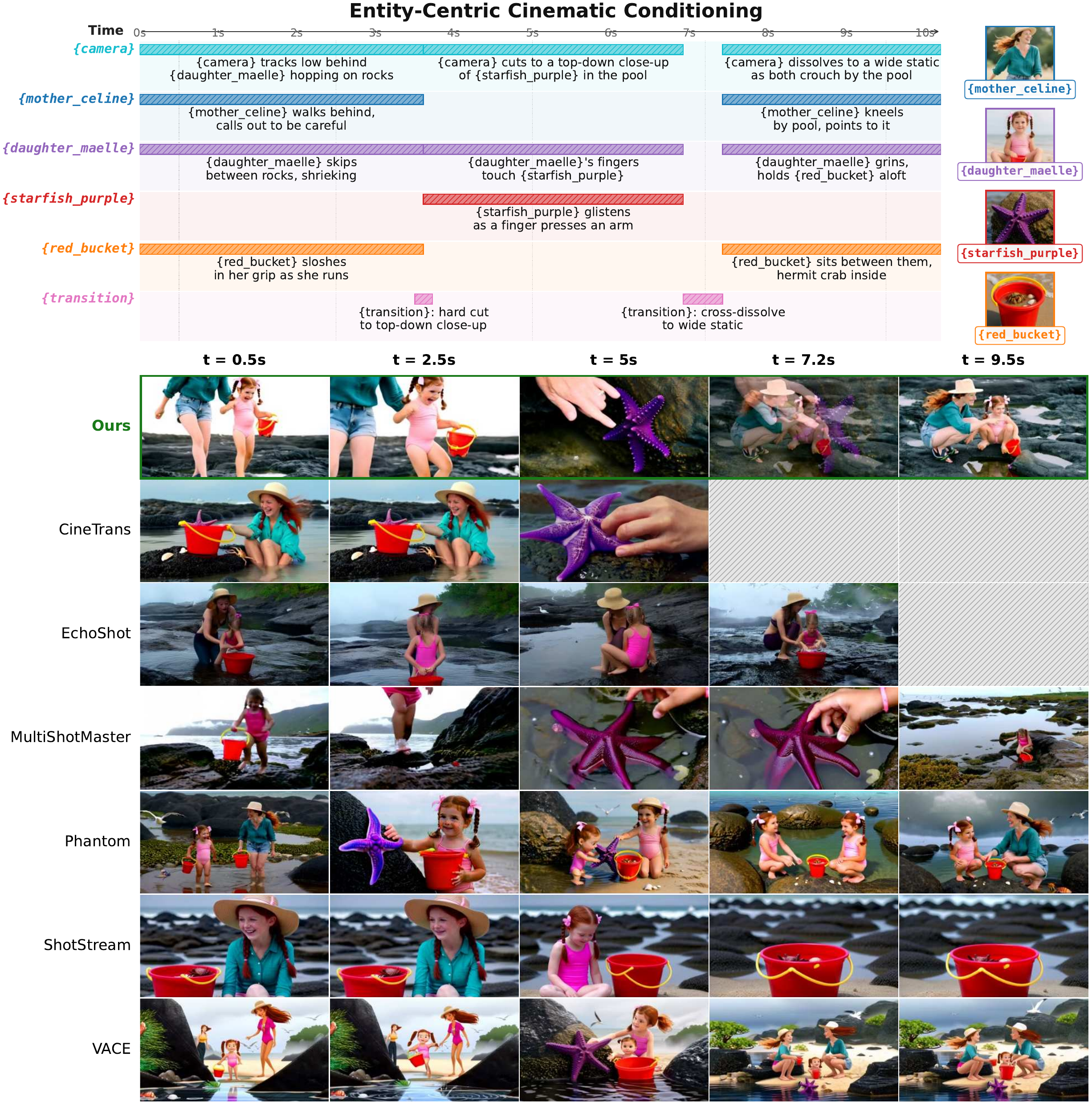}
    \mycaption{Additional qualitative comparison on \ourbenchsyn}{
    }
    \label{fig:qualitative_extra_syn_3}
\end{figure*}

\begin{figure*}[t]
    \centering
    \includegraphics[width=\textwidth]{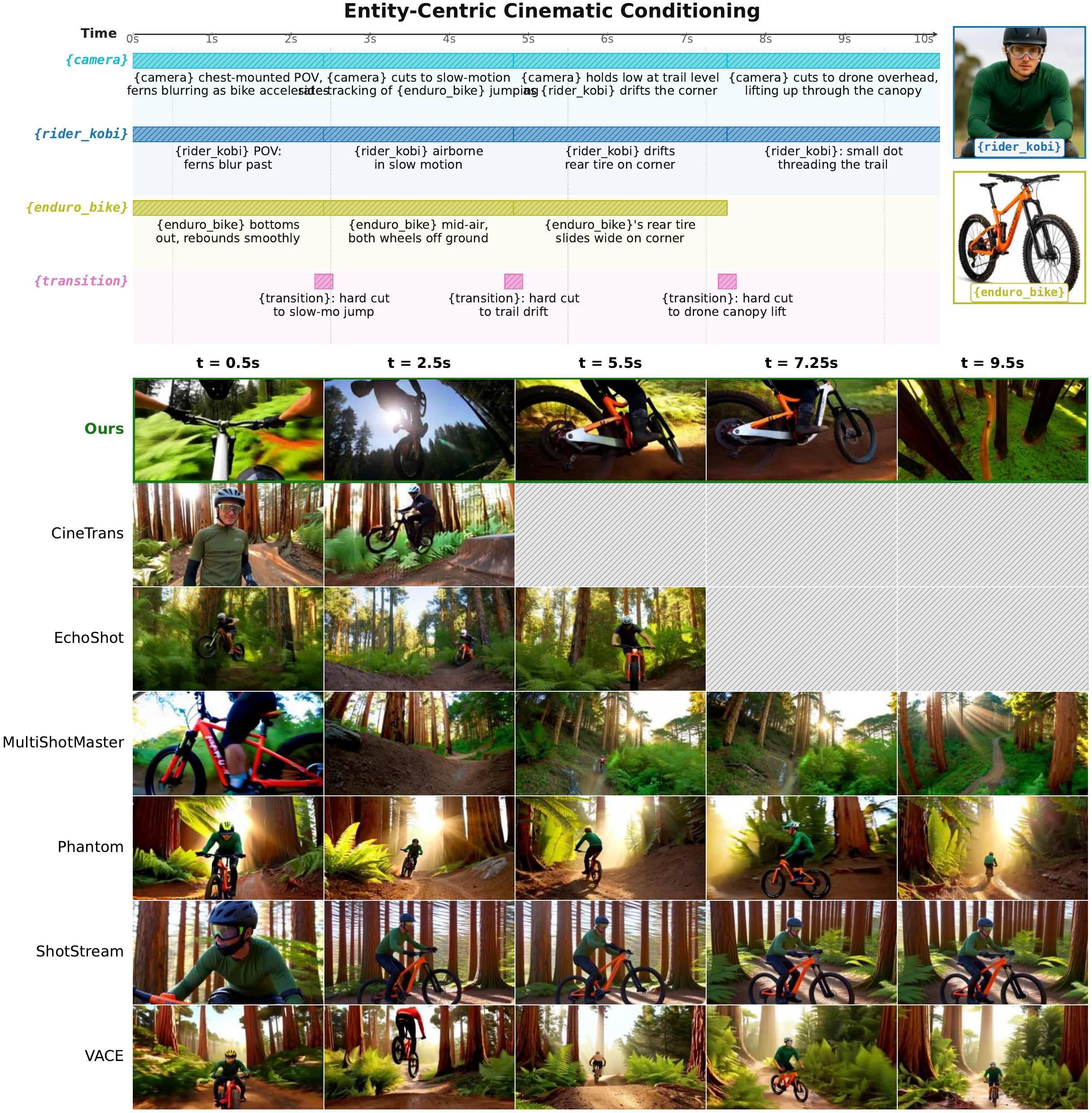}
    \mycaption{Additional qualitative comparison on \ourbenchsyn}{
    }
    \label{fig:qualitative_extra_syn_4}
\end{figure*}

We report the quantitative evaluation on \ourbenchsyn in \cref{tab:quantitative_syn}, and provide additional qualitative comparisons at the default resolution $288{\times}512$ in \cref{fig:qualitative_extra_real_1,fig:qualitative_extra_real_2} (\ourbench) and \cref{fig:qualitative_extra_syn_1,fig:qualitative_extra_syn_3,fig:qualitative_extra_syn_4} (\ourbenchsyn). Please refer to our project website for video comparisons and additional results.

\begin{figure*}[t]
    \centering
    \includegraphics[width=\textwidth]{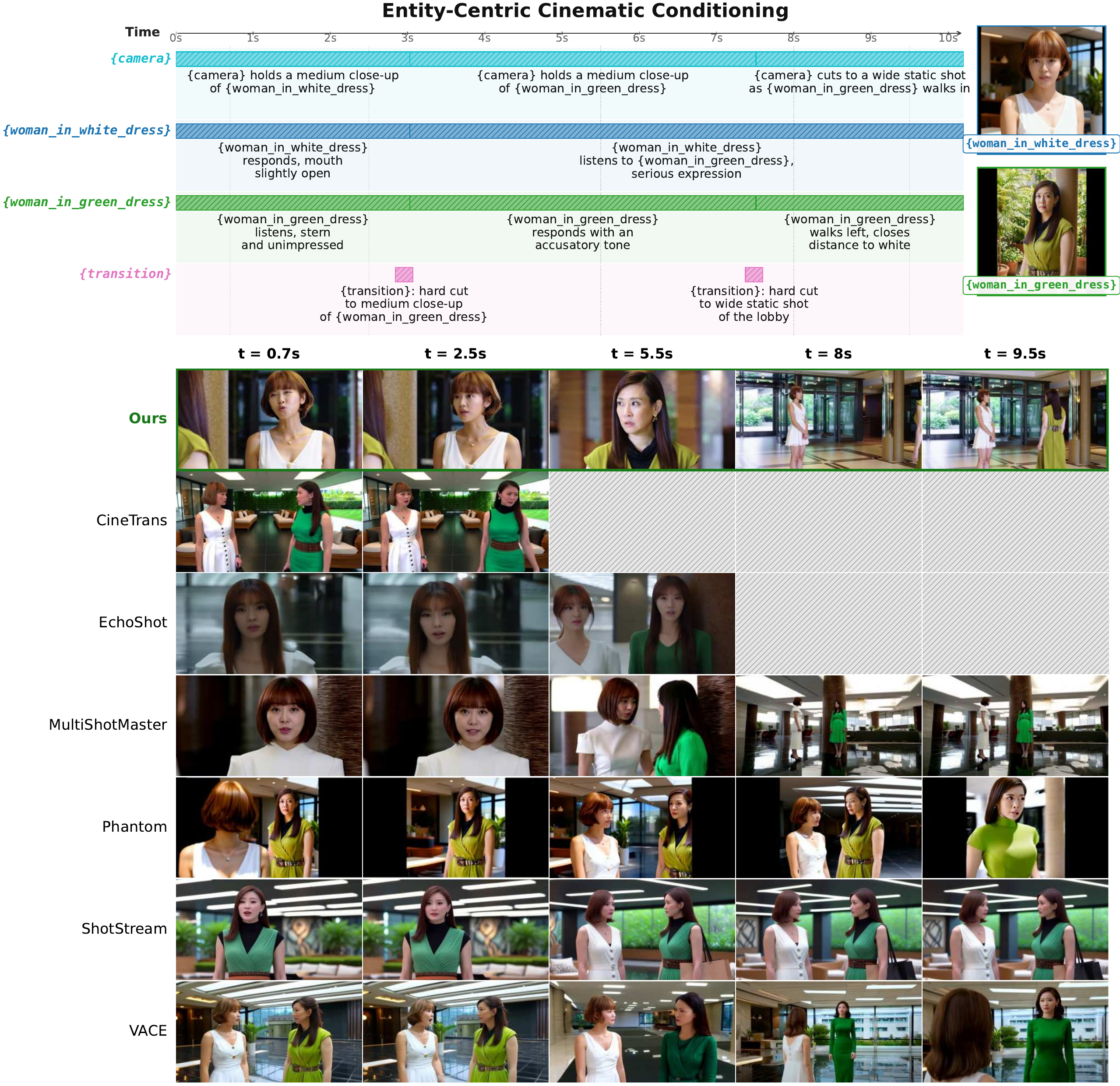}
    \mycaption{Qualitative comparison at $720\mathrm{p}$ resolution for \ourmodel on \ourbench}{
    }
    \label{fig:qualitative_extra_720p_real}
\end{figure*}

\begin{figure*}[t]
    \centering
    \includegraphics[width=\textwidth]{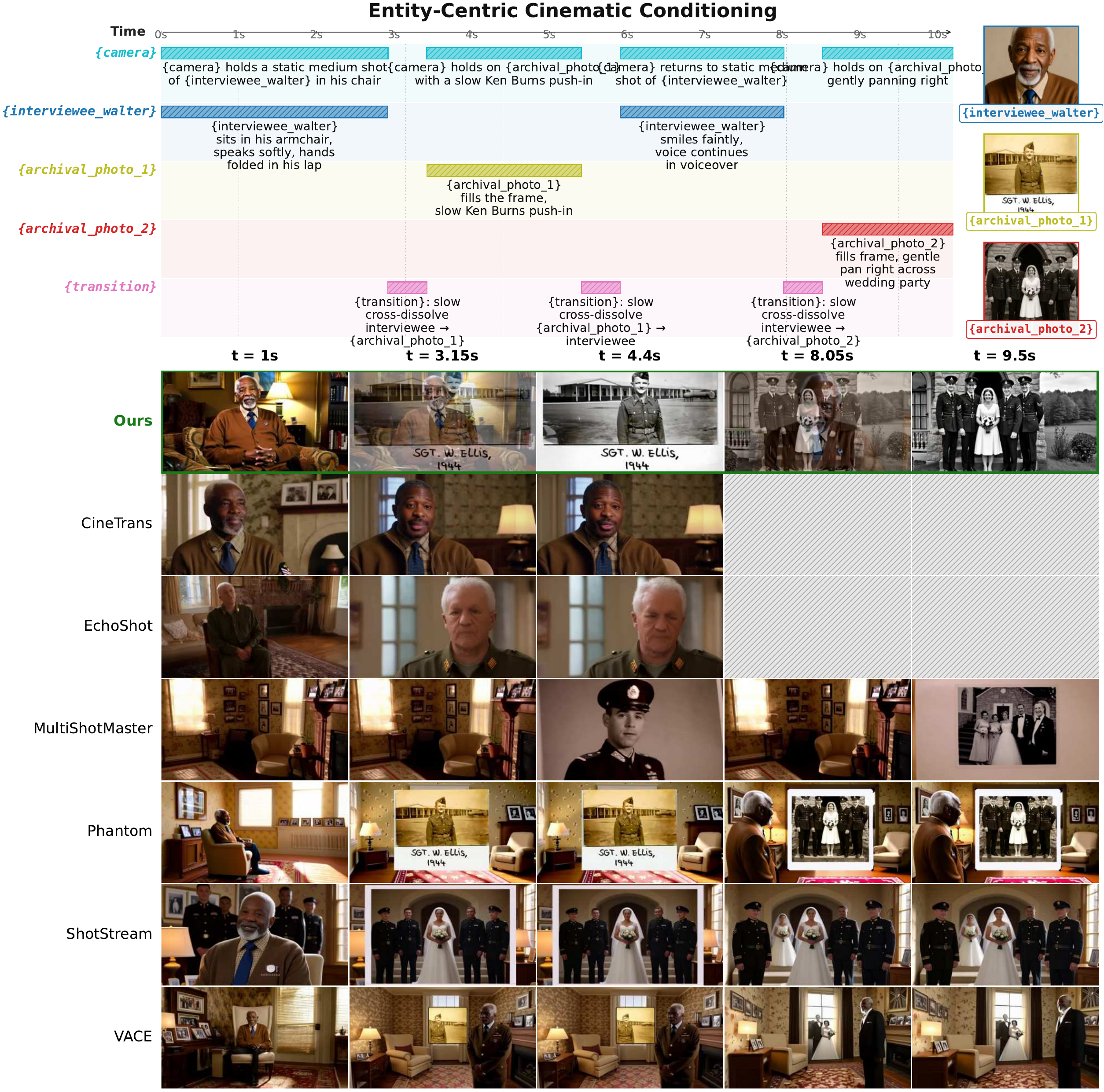}
    \mycaption{Qualitative comparison at $720\mathrm{p}$ resolution for \ourmodel on \ourbenchsyn}{
    }
    \label{fig:qualitative_extra_720p_syn}
\end{figure*}

\cref{fig:qualitative_extra_720p_real,fig:qualitative_extra_720p_syn} report comparisons at $720{\times}1280$ for 10s long videos even when our model is trained only up to 5 seconds. 
The four control axes the framework targets transfer cleanly to the higher resolution: per-subject reference identity is preserved across shot boundaries, camera primitives match their textual specification, shot transitions occur at the prescribed timestamps with the prescribed type (hard cut, cross-dissolve, wipe), and dense per-event captions are reflected in the relevant frames. By generating more tokens at $720\text{p}$, the model achieves much higher quality, characterized by richer detail, enhanced aesthetic appeal, and smoother motion dynamics compared to lower-resolution baselines. 

\begin{figure*}[t]
    \centering
    \includegraphics[width=\textwidth]{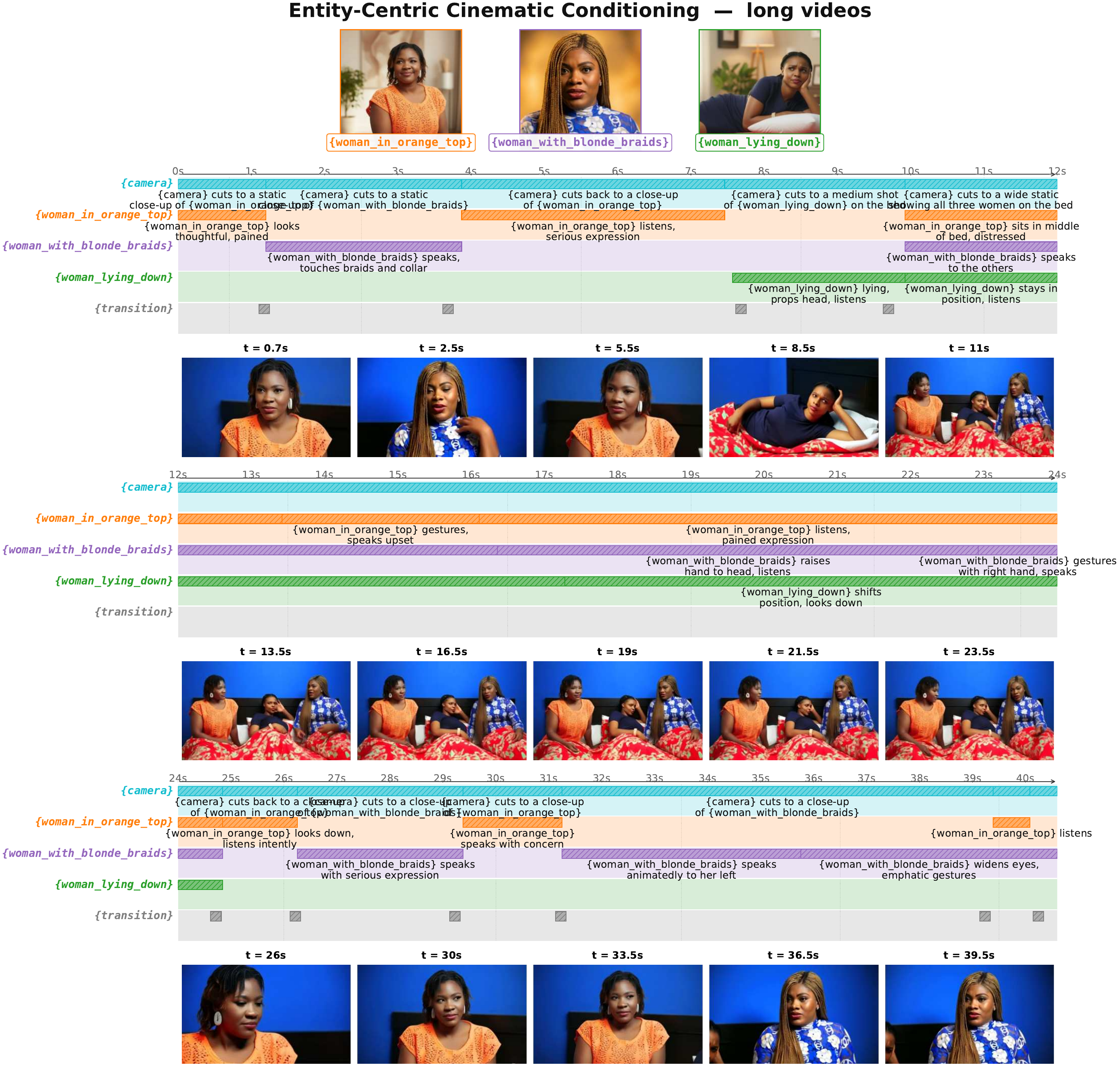}
    \mycaption{Long video generation (40s) from \ourmodel on \ourbench}{
    }
    \label{fig:qualitative_long_1}
\end{figure*}

\begin{figure*}[t]
    \centering
    \includegraphics[width=\textwidth]{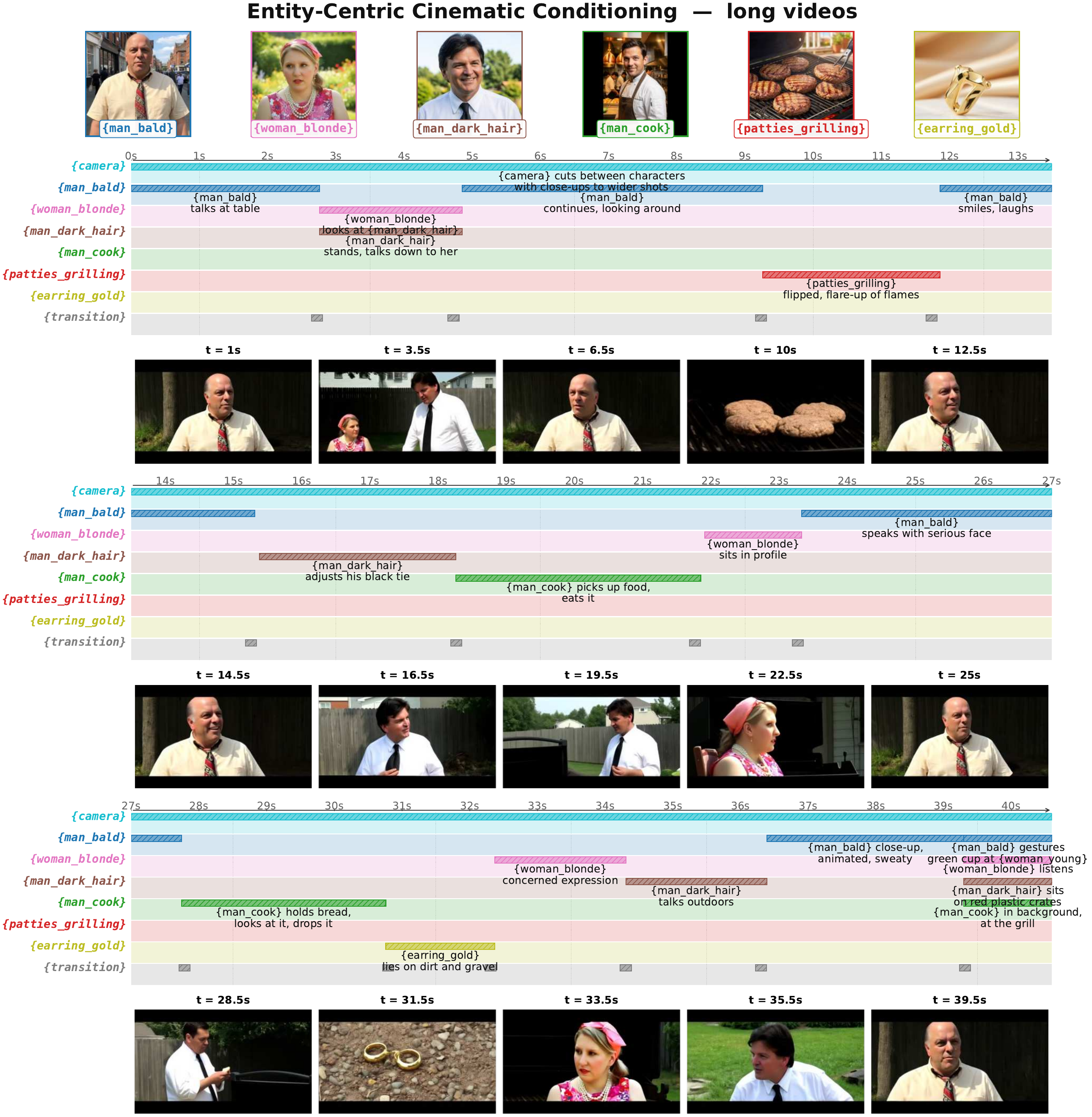}
    \mycaption{Long video generation (40s) from \ourmodel on \ourbench}{
    }
    \label{fig:qualitative_long_2}
\end{figure*}

\cref{fig:qualitative_long_1,fig:qualitative_long_2} extend the evaluation to 40s clips with substantially more scripted shot transitions and dense, time-localized events spanning the full duration. 
Identity, scene grounding, and dense-caption alignment remain stable over the longer horizon; entities reappear consistently after intervening shots, camera primitives resolve to the requested framings, and shot transitions trigger at the right moments with the right type. We observe minimal qualitative degradation maintaining high levels of entity caption reference binding as the duration is scaled from the 10s training horizon to 40s at inference.

\begin{table}[t]
    \centering
    \mycaption{Quantitative comparison of cinematic conditioning on \ourbenchsyn}{
        Best in \textbf{bold}, second-best \underline{underlined}.
    }
    \vspace{+2mm}
    \label{tab:quantitative_syn}
    \footnotesize
    \setlength{\tabcolsep}{3.5pt}
    \begin{tabular}{l|c|c|cccc|c}
        \toprule
        & \multicolumn{1}{c|}{\textbf{Subject ID}} & \multicolumn{1}{c|}{\textbf{Global Caption}}& \multicolumn{4}{c|}{\textbf{Dense Caption Following} (ViCLIP$\uparrow$)}  & \textbf{Transition Timing}  \\
        \cmidrule(lr){2-2}\cmidrule(lr){3-3}\cmidrule(lr){4-7}\cmidrule(lr){8-8}
        \textbf{Method}                        & DINO$\uparrow$ & CLIP$\uparrow$ & Subject & Scene & Camera & Transition & Recall$\uparrow$ \\
        \midrule
        Phantom~\cite{phantom}                   & \textbf{0.601} & \underline{0.311} & 0.237 & \underline{0.240} & 0.215 & 0.125 & 0.114 \\
        VACE~\cite{vace}                         & \underline{0.562} & \textbf{0.315} & 0.239 & \textbf{0.243} & 0.220 & 0.128 & 0.115 \\
        \midrule
        CineTrans~\cite{cinetrans}               & 0.489 & 0.295 & \underline{0.241} & 0.227 & \underline{0.232} & \underline{0.145} & 0.230 \\
        EchoShot~\cite{echoshot}                 & 0.371 & 0.283 & 0.224 & 0.226 & 0.205 & 0.124 & 0.053 \\
        MultiShotMaster~\cite{multishotmaster}   & 0.426 & 0.290 & 0.239 & 0.238 & 0.231 & 0.140 & \underline{0.258} \\
        ShotStream~\cite{shotstream}             & 0.466 & 0.290 & 0.224 & 0.206 & 0.206 & 0.132 & 0.165 \\
        \midrule
        \rowcolor{gray!12}
        \textbf{\ourmodel (ours)}                & 0.556 & 0.310 & \textbf{0.245} & \underline{0.240} & \textbf{0.251} & \textbf{0.146} & \textbf{0.360} \\
        \bottomrule
    \end{tabular}
\end{table}





\end{document}